%% file: neurips_2026.tex
\documentclass{article}
\PassOptionsToPackage{numbers, compress}{natbib}
\usepackage[main,final]{neurips_2026}
\usepackage[utf8]{inputenc}
\usepackage[T1]{fontenc}
\usepackage{nicefrac}
\usepackage{fontawesome5}

\input{math_commands.tex}

\usepackage{hyperref}
\usepackage{url}
\usepackage{wrapfig}
\usepackage{microtype}
\usepackage{float}
\usepackage{graphicx}
\usepackage{booktabs}
\usepackage{subcaption}
\usepackage{amsmath,amssymb,mathtools,amsthm}
\usepackage{enumitem}
\usepackage{tabularx,adjustbox,placeins}
\usepackage{dblfloatfix}
\usepackage{bbm}
\usepackage{booktabs}
\usepackage{caption}
\usepackage{longtable}
\usepackage{array}
\usepackage{xcolor}
\usepackage{tcolorbox}
\definecolor{softredB}{RGB}{255, 160, 160}

\newcommand\blfootnote[1]{%
  \begingroup
  \renewcommand\thefootnote{}\footnote{#1}%
  \addtocounter{footnote}{-1}%
  \endgroup
}

\newtcolorbox{PromptBlock}{
  colback=gray!5,
  colframe=black!40,
  boxrule=0.4pt,
  arc=2pt,
  left=6pt,
  right=6pt,
  top=4pt,
  bottom=4pt
}
\definecolor{softblue}{RGB}{150, 180, 255}

\title{Failing to Falsify: Evaluating and Mitigating Confirmation Bias in Language Models}

\author{
Ayush Rajesh Jhaveri$^{1,2}$,
Anthony GX-Chen$^{1}$,
Ilia Sucholutsky$^{1}$,
Eunsol Choi$^{1}$ \\
$^{1}$New York University \\
$^{2}$Capital One\thanks{This research was completed prior to joining Capital One.} \\
\texttt{\{ayush.rajesh.jhaveri,eunsol\}@nyu.edu} 
}

\definecolor{guesscolor}{HTML}{A9C8F2}
\definecolor{testcolor}{HTML}{F4A9A3}
\definecolor{feedbackcolor}{HTML}{FFEB70}
\begin{document}

\maketitle

\begin{abstract}
Confirmation bias, the tendency to seek evidence that supports rather than challenges one's belief, hinders one's reasoning ability. We examine whether large language models (LLMs) exhibit confirmation bias by adapting the rule-discovery study from human psychology: given a sequence of three numbers (a ``triple''), an agent engages in an interactive feedback loop where it (1) proposes a new triple, (2) receives feedback on whether it satisfies the hidden rule, and (3) guesses the rule.
Across eleven LLMs of multiple families and scales, we find that LLMs exhibit confirmation bias, often proposing triples to confirm their hypothesis rather than trying to falsify it. This leads to a slower and less frequent discovery of the hidden rule. We further explore intervention strategies developed for humans (e.g., encouraging the agent to consider counterexamples). We find that prompting LLMs with such instructions consistently decreases confirmation bias in LLMs, improving rule discovery rates from 42\% to 56\% on average. Lastly, we mitigate confirmation bias by distilling intervention-induced behavior into LLMs, showing generalization to a new domain (the Blicket test) and beyond rule discovery (mathematical self-correction). Our work shows that confirmation bias is a limitation of LLMs in hypothesis exploration, and that it can be mitigated via injecting interventions designed for humans.\blfootnote{You can find data and code for this paper here: \url{https://github.com/ayushjhaveri/llm-confirmation-bias}} 
\end{abstract}

\section{Introduction}
\label{submission}

\begin{figure}[h]
  \centering
  \vspace{-5pt}
  \includegraphics[width=\textwidth]{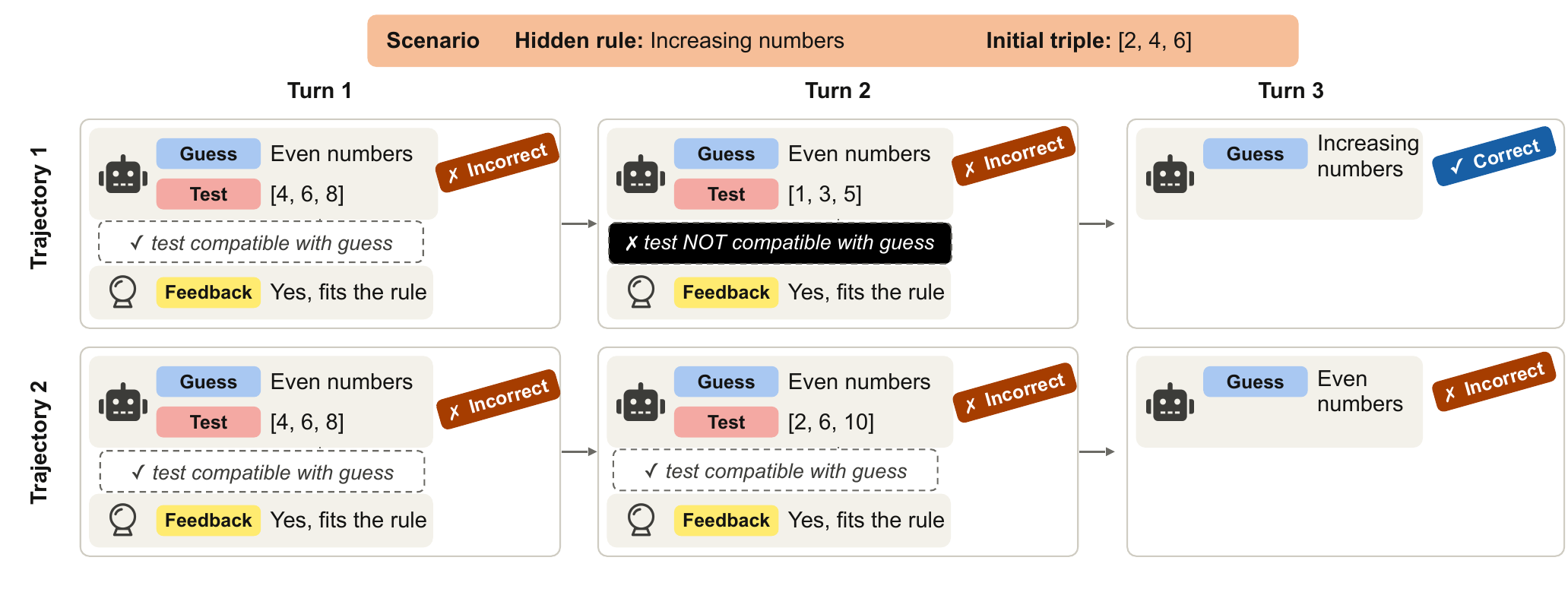}
  \vspace{-10pt}

\caption{\textbf{Experimental Setting to Measure Confirmation Bias.}
We show two trajectories for a rule discovery task, in which an agent aims to
infer a hidden rule over multiple turns. Starting from an initial
triple, the agent (\smash{\faRobot{}}) states a \smash{\colorbox{guesscolor}{guess}} about the rule and
proposes a new triple to \smash{\colorbox{testcolor}{test}}, then receives binary
\colorbox{feedbackcolor}{feedback} on whether that triple satisfies the hidden
rule. A \textbf{compatible test} is consistent with the agent's current guess,
whereas an \textbf{incompatible test} contradicts it. Trajectory~1 introduces an incompatible test in
the second turn, which eliminates the incorrect guess and leads to discovery
of the correct rule. In contrast, Trajectory~2 proposes
compatible tests in both turns, showing confirmation bias, and keeps the
incorrect guess. }
  
  
  \vspace{5pt}
  \label{fig:schematic}
\end{figure}

Humans often form an initial hypothesis and then seek evidence that supports their hypothesis rather than refutes it~\citep{Nickerson1998ConfirmationBA}.
For example, a doctor who suspects a certain diagnosis may focus on specific symptoms that support it and interpret positive test results as confirmation. 
Such selective exploration, known as \textbf{confirmation bias}, leads to inefficient learning and decision-making. Figure~\ref{fig:schematic} exemplifies such a scenario, where consistently using confirmatory examples leads to failed rule discovery~\citep{wason1960failure}.

We aim to understand confirmation bias for LLMs acting as agents~\citep{yao2022react, wang2024survey, shinn2023reflexion}. In these scenarios, LLMs perform tasks that require exploratory reasoning: generating hypotheses, testing them, and refining beliefs in iterative loops. An agent that only verifies its current guess will be suboptimal compared to an agent which also seeks evidence that can falsify its current guess. To this end, we adapt a simplified rule discovery study in humans~\citep{wason1960failure} and synthetically generate large-scale data, where an initial triple can match multiple valid rules. We then evaluate the LLMs' ability to identify the hidden rule by proposing and testing new triples. 
This setup enables step-by-step observation of how models revise hypotheses based on observations. We further quantify a model's confirmation bias by measuring the ratio between incompatible and compatible tests it proposes.

We experiment with eleven LLMs of different families and scales. Models exhibit varying extents of confirmation bias and task success rates (from 6\% to 78\%, within a maximum turn limit of 45 interactions). We find recent LLMs with longer reasoning traces~\citep{qwq32b,yang2025qwen3} show much stronger performance, as well as lower confirmation bias. As was in humans, we observe a negative correlation between the extent of confirmation bias and task success rates. 

We then study methods to decrease confirmation bias, using two well-established interventions originally developed for humans: Dual-Goal~\citep{gale2006} and Think-in-Opposites~\citep{branchini2023}. Prompting LLMs with these strategies leads to meaningful performance gains. We then investigate whether fine-tuning LLMs via symbolic knowledge distillation~\citep{west2022symbolicknowledgedistillationgeneral} to use these intervention strategies helps decrease confirmation bias, alleviating the need for task-specific inference-time interventions. Finally, we study whether the fine-tuned models generalize to a new rule-discovery domain, the Blicket Test~\citep{gopnik2000detecting}, and beyond rule discovery to mathematical reasoning~\citep{opencompass2025aime,bytedance_seed_2025_beyondaime}. On the Blicket Test, the fine-tuned models exhibit lower confirmation bias and improved task success, while on mathematical reasoning tasks, they show improved self-correction after incorrect solutions.

Together, our framework enables studying whether LLMs exhibit confirmation bias during hypothesis exploration. Intervention studies further present promising results in how psychologically inspired framings can reduce this bias and improve task performance. By internalizing debiasing behaviors through symbolic knowledge distillation, we show that these behaviors can generalize across rule-discovery domains and beyond rule discovery to broader reasoning tasks.

\section{Background}
\label{sec:design}
We first describe Wason's seminal 1960 study which quantified the notion of confirmation bias. 
 \paragraph{Setting} Human participants were shown the triple of integers (i.e., \texttt{2-4-6}), told it conformed to a hidden rule (which was any three numbers in increasing order), and asked to discover that rule by proposing new triples.  
After each proposal, an experimenter responded \texttt{YES} or \texttt{NO}, indicating whether the proposed triple conformed to the hidden rule. Participants could guess the rule at any time; incorrect guesses did not terminate the experiment. The experiment continued until the participant correctly guessed the rule, expressed a wish to give up, or the time limit (45 minutes) was reached. 

\paragraph{Measuring Confirmation Bias} Confirmation bias was quantified by measuring the ratio between the number of turns when a participant tests triples that contradict (\emph{incompatible}) versus confirm (\emph{compatible}) their current hypothesis. Participants were required to record the reason for each proposed triple, which were used to infer their current hypothesis. Higher ratios indicate more disconfirmatory testing and thus lower confirmation bias.

\paragraph{Results}
The study included 29 participants. $6/29$ participants announced the correct rule on their {first} attempt, and $21/29$ eventually discovered the correct rule. Participants who were correct on their first announcement exhibited more
disconfirmatory exploration, with a mean I:C of $1.79$, compared to $0.24$ for those whose first announcement was incorrect.
Among who initially failed, those who succeeded on the \emph{second} announcement showed a higher I:C ($0.50$) than those who did not ($0.19$). These results show that in humans, higher I:C correlates with earlier and more frequent rule discovery.

\paragraph{Caveats} Later follow up work~\cite{Klayman1987ConfirmationDA} refined Wason study, arguing positive testing strategy does not necessarily mean confirmation bias. Compatible test can falsify the current hypothesis if the hidden hypothesis is narrower than the current hypothesis. In this study, we follow Wason's study design, as target hypotheses are generally broad. 

\noindent\textbf{Studies of Wason 2-4-6 in LLMs} 
\citet{o2025confirmation} conducts a small-scale Wason 2-4-6 study using two-guess sessions, making qualitative judgments based on similarity between successive guesses. They do not evaluate exploration through test-triple selection or consider multiple rules and instances. Another study~\citep{banatt2024wilt} adapted Wason 2-4-6 to evaluate multi-turn reasoning in LLMs. Here, each episode initializes a hidden Boolean rule over three numbers, and the model may propose up to 30 test triples, receiving binary feedback before making a rule guess. They report the guessing accuracy, framing it as a small-scale evaluation benchmark (N=50) for reasoning and hypothesis exploration abilities for LLMs without focusing specifically on confirmation bias. The evaluated LLMs achieve at most a 26\% task success rate. We expand this approach substantially by introducing a larger-scale dataset and a framework to measure the confirmation bias. We use this framework to study mitigation strategies for reducing LLM confirmation bias and improving performance. 





\vspace{-0.3em}
\section{Framework to Identify Confirmation Bias in LLMs }
\label{sec:framework}
\vspace{-0.3em}

\begin{tcolorbox}[
  colback=gray!5,
  colframe=black!40,
  boxrule=0.4pt,
  arc=2pt,
  left=6pt, right=6pt, top=4pt, bottom=4pt,
  title={\textbf{Notation Summary}},
  fonttitle=\small
]
\small
\begin{tabular}{@{}llp{8cm}@{}}
 Test Scenario & ($\mathbf{x}_{\text{ini}}$ ,$\mathbf{r}^*$) &   (Initial triple of integers (input), Hidden target rule) \\
 Episode & $e$ & A full sequence of rule discovery game: initial triple followed by 45 guess, test, feedbacks turns \\
 Guess &$\mathbf{r}_i$ &  Model's guessed rule at turn $i$ \\
 Test  &$\mathbf{x}_i$&  Triple of integers proposed at test turn $i$ \\
 Feedback & $y_i=\mathbf{r_i}(\mathbf{x_i})$ & A binary label whether triple $\mathbf{x_i}$ conforms to rule $\mathbf{r_i}$ \\
 First correct turn & $t_e^\star$ &  Turn index of the first correct rule announcement in episode $e$, if any
  \\
\end{tabular}
\end{tcolorbox}
 \paragraph{Setting}
We simulate a multi-turn reasoning game for LLMs. Each game scenario consists of an initial triple of integers ($\mathbf{x}_{\text{ini}}$) and a hidden rule $\mathbf{r}^*$. Initially, the LLM is given $\mathbf{x}_{\text{ini}}$ and the task instruction.\footnote{The full task instruction can be found in Appendix~\ref{appendix:prompts}.}
Then, LLM takes alternating \texttt{Guess} and \texttt{Test} turns. On $i$-th \texttt{Guess} turn, the LLM states the guessed hidden rule $\mathbf{r}_{i}$ in text; on the subsequent ($i$-th) \texttt{Test} turn, it proposes a new triple $\mathbf{x}_i$ for evaluation. We denote that a triple $\mathbf{x}$ conforms to a rule $\mathbf{r}$ as $\mathbf{r}(\mathbf{x}) = 1$, and $\mathbf{r}(\mathbf{x}) = 0$ otherwise. 

Then, the environment provides binary feedback $y_i$ whether the proposed triple conforms to the hidden rule ($\mathbf{r_i}(\mathbf{x_i})$). We continue this for $45$ turns. For episodes in which the model correctly identifies the hidden rule, let $t_e^\star$ denote the turn index of the first correct announcement, i.e., $Eq(\mathbf{r}_{t_e^\star}, \mathbf{r}^*) = 1$. 
We describe how our setting differs from Wason study on humans in the appendix~\ref{appendix:divergencewason}.

\paragraph{Evaluation Metrics}\label{subsec:eval_metrics}
We report reasoning process success, efficiency of the reasoning process, as well as the extent of confirmation bias exhibited during the multi-turn games. 

\noindent\textbf{(1) Task Success (Success ratio \% $\uparrow$)} We report the percentage of scenarios in which the LLM guesses the correct rule during the conversation. For each episode, we measure whether any of $i$-th guessed rule $\mathbf{r}_i$ matches the hidden rule $\mathbf{r}^*$ using LLM-as-a-judge for each scenario, $\mathbf{1}\Big( \bigvee_{i=1}^{n} Eq(\mathbf{r}_i, \mathbf{r}^*) \Big)$, where $n$ is the maximum number of turns.

\noindent\textbf{(2) \# Turns until Success ($\downarrow$)}
We report the number of turns LLMs took to guess the correct rule. This is only computed for episodes where the model successfully guessed the correct rule within the allowed maximum number of turns. More formally, 
$\frac{1}{|\mathcal{E}_{\mathrm{succ}}|}
\sum_{e \in \mathcal{E}_{\mathrm{succ}}}
t_e^*$ where $\mathcal{E}_{\mathrm{succ}}$ denotes the set of episodes in which the model correctly identifies the hidden rule.

\noindent\textbf{(3) \# Tokens per Turn ($\downarrow$)}
We report the average number of tokens generated by the model per \texttt{Guess} or \texttt{Test} turn across all episodes. This metric serves as a proxy for the amount of compute spent on reasoning
and hypothesis testing.
At comparable task performance, lower values indicate more concise and efficient
reasoning.

\noindent\textbf{(4) Confirmation Bias Status (Incompatible:Compatible Ratio (I:C) $\uparrow$).} Following \citet{wason1960failure}, we quantify the model’s confirmation bias by measuring the ratio between the number of turns when it tests triples that contradict (\emph{incompatible}) versus confirm (\emph{compatible}) its current hypothesis.

For episode $e$, let $C_e$ and $I_e$ denote the numbers of
\emph{compatible} and \emph{incompatible} tests, respectively:
\[
C_e = \sum_{t=1}^{t_e^\star} \mathbbm{1}\!\left[\mathbf{r_t(x_t)}=1\right], \quad
I_e = \sum_{t=1}^{t_e^\star} \mathbbm{1}\!\left[\mathbf{r_t(x_t)}=0\right], \quad
\mathrm{I{:}C} = \frac{\sum_{e \in \mathcal{E}} I_e}{\sum_{e \in \mathcal{E}} C_e}
\]
where $\mathcal{E}$ denotes the total set of evaluation episodes.

To compute the compatibility of a triple with a rule, we first use an LLM to translate each rule into a Python function. Then we execute this Python function to decide the compatibility between the rule and the triple. The implementation details are in Appendix~\ref{appendix:compatibility-judge}. 

\begin{table}
  \caption{Example rule sets (each set has 4 rules) and sampled triples. Each rule is defined over integer triples $(a,b,c)$. All triples are sampled from the feasible intersection of the rules.}
  \label{tab:example-rule-group}
  \vspace{-0.5em}
  \begin{center}
    \begin{small}
      \begin{tabular}{p{0.6\columnwidth}p{0.33\columnwidth}}
        \toprule
        Rule group & Initial triple samples \\  
        \midrule
        All odd; $b$ is (strict) max; mixed signs; at least one multiple of 3
        &
        $(-9,\,55,\,-71)$,\ $(-77,\,75,\,-47)$ \\
        All prime numbers; non-increasing; all positive; contains a prime
        &
        $(79,\,43,\,31)$,\ $(89,\,67,\,2)$ \\
        \bottomrule
      \end{tabular}
    \end{small}
  \end{center}
\end{table}

\paragraph{Data}

Each game scenario requires an initial triple $\mathbf{x}_{\text{ini}}$, the target rule $\mathbf{r}^*$, and a set of N other rules $\{\mathbf{r}_i\}_{i=1}^{N}$ that satisfies the initial triple. We ensure there are other plausible rules that match the initial triple so that the LLM tests other triples to identify the hidden rule. We aim for a diverse, plausible rule set. Table~\ref{tab:example-rule-group} contains example rule sets and sampled initial triples.

We develop 4 rule sets for training, 2 for validation, and 4 for testing, with each rule set containing 4 rules. For each rule set, we sample triples that satisfy all four rules, and evaluate each triple against each rule as the target, yielding one episode per triple–rule pairing. Training uses 100 triples per rule set (1,600 evaluation episodes), validation uses 2 triples per rule set (16 episodes), and testing uses 5 triples per rule set (80 episodes). All reported evaluation metrics are computed on the held-out test split unless otherwise specified. The training and validation splits are used for distillation (Section~\ref{sec:distillation}). We describe the data generation process below.

\noindent\textbf{Rule Set Generation} We prompt an LLM to generate 40 game scenarios. First, we prompt the LLM to generate ten sets of four plausible rules over integer triples \((a,b,c)\in[-99,100]^3\). The prompt encourages rules from several conceptual families: \emph{ordering}, \emph{arithmetic structure}, \emph{parity or divisibility}, \emph{sign}, and \emph{number properties}. We instruct that there should be non-empty set of triples that can satisfy all four rules. After initial generation, we apply heuristic checks (e.g., detecting multi-step arithmetic dependencies or rules relying on a single variable). Rule sets that fail these checks are revised via reprompting, resulting in approximately 33\% of the initial generations being corrected. 

After filtering, we replace one of the LLM-generated rules in each rule set with a human-derived rule from prior work~\citep{DVN/A8ZWLF_2018,Tenenbaum2000RulesSimilarity}.
The replacement is chosen such that there remains at least one triple satisfying all four rules.
The data details are in Appendix~\ref{appendix:data}.

\vspace{-0.4em}
\section{Replicating Human Studies}
\label{sec:exp-setup}
We now evaluate off-the-shelf LLMs with three interaction settings: Baseline, Dual-Goal, and Think-in-Opposites. The first simulates the Wason's study, and the other two mimic two intervention studies respectively. We ask the following two research questions: (1) Do LLMs exhibit confirmation bias? (2) Are intervention strategies to reduce the confirmation bias in humans effective for LLMs?

\subsection{Interaction Settings}
\label{sec:variants}

\noindent\textbf{Baseline} This adapts the instruction from Wason’s classic 2--4--6 rule-discovery task~\citep{wason1960failure} as is. The prompt and example interaction in Appendix~\ref{appendix:baseline-prompt}. 
Each episode begins with an initial \texttt{Guess} turn, where the model states its first guess at the rule. It continues with the model's \texttt{Test}, feedback on the test from the environment, then a new \texttt{Guess} turn.

\noindent\textbf{Think-in-Opposites (TiO)}
This adapts the study of \citet{branchini2023} and instructs the LLM to focus on one property of the previous \textbf{Test} example triple and construct a new triple that is opposite with respect to that property. The prompt and example interaction for this intervention is in Appendix~\ref{appendix:opposites-prompt}. Except for the prompt, it follows the same interaction protocol as the baseline condition. 

\noindent\textbf{Dual-Goal} We introduce a prompt adapted from the Dual-Goal paradigm~\citep{gale2006} (prompt and example in Appendix~\ref{appendix:dualgoal-prompt}).
The interaction protocol is identical to the baseline condition, with two modifications. First, during the \texttt{Guess} turn, the LLM must output two test rules: one for the DAX rule and one for the MED rule. The \texttt{DAX} rule corresponds to the original hidden rule, while the \texttt{MED} rule is its logical complement, capturing triples that violate the \texttt{DAX} rule. Second, during the \texttt{Test} turn, the LLM proposes a triple. The environment will evaluate this the same way (whether it fits the hidden rule or not), but instead of \texttt{YES}/\texttt{NO} feedback, it will output \texttt{DAX}/\texttt{MED} respectively.

\subsection{Models Evaluated}
\label{sec:model-setup}

We evaluate 11 LLMs, including those with thinking mode, which generate intermediate reasoning traces, and LLMs without it. Decoding parameters, post-processing rules, and judge prompts are in
Appendix~\ref{appendix:judge} and we manually validate LLM-as-a-judge performance
in Appendix~\ref{appendix:judge-validation}.

\textbf{LLMs (standard inference)}
Qwen3-8B, Qwen3-14B, and Qwen3-32B~\citep{yang2025qwen3};
Llama-3.3-70B-Instruct~\citep{llama3_3_70b};
and GPT-4o~\citep{openai_gpt4o}. For Qwen3 models, we disable thinking by setting \texttt{enable\_thinking=False} in the model's chat template.

\textbf{LLMs with Thinking Mode}
Gemini-2.5-Pro~\citep{comanici2025gemini}; Qwen3-8B, Qwen3-14B, and Qwen3-32B~\citep{yang2025qwen3};
QwQ-32B~\citep{qwq32b};
and DeepSeek-R1-Distill-Llama-70B~\citep{deepseek_r1}.
Thinking-mode models generate reasoning traces before an answer. For Qwen3 models, we enable thinking by setting \texttt{enable\_thinking=True} in the model's chat template. Gemini-2.5-Pro, QwQ-32B, and DeepSeek-R1-Distill-Llama-70B generate reasoning as part of their default inference behavior.

\subsection{Results}
\label{sec:results}

\begin{table*}
\centering
\small
\setlength{\tabcolsep}{2.5pt}
\renewcommand{\arraystretch}{0.8}
\caption{\textbf{LLMs task performance and the extent of confirmation bias (I:C ratio) on Wason's rule discovery game.} We observe varying success rates for LLMs. Overall, larger models with thinking mode show higher task success rates. TiO represents the Think-in-Opposites intervention. }
\label{tab:replication}
\begin{tabular}{l| c c c c | c c c c}
\toprule
& \multicolumn{4}{c|}{\textbf{Standard Inference}} &
  \multicolumn{4}{c}{\textbf{Thinking Mode Inference}} \\
\cmidrule(lr){2-5} \cmidrule(lr){6-9}
& \textbf{Task Suc.} &
  \textbf{Avg. \# Turns} &
  \textbf{Avg. \# Tkns} &
  \textbf{I:C} &
  \textbf{Task Suc.} &
  \textbf{Avg. \# Turns} &
  \textbf{Avg. \# Tkns} &
  \textbf{I:C} \\
& \textbf{\% $\uparrow$} &
  \textbf{until Suc. $\downarrow$} &
  \textbf{per Turn $\downarrow$} &
  \textbf{$\uparrow$} &
  \textbf{\% $\uparrow$} &
  \textbf{until Suc. $\downarrow$} &
  \textbf{per Turn $\downarrow$} &
  \textbf{$\uparrow$} \\
\midrule

& \multicolumn{4}{c|}{\textbf{Qwen3-8B}} &
  \multicolumn{4}{c}{\textbf{Qwen3-8B}} \\
Baseline
& 0.06 & 5.00 & 8.4 & 0.35
& 0.21 & 10.88 & 1088.2 & 0.15 \\
TiO
& 0.06 & 15.25 & 8.4 & 0.30
& 0.36 & 9.01 & 1716.1 & 0.35 \\
Dual-Goal
& 0.08 & 0.00 & 13.4 & 0.57
& 0.40 & 7.37 & 2142.7 & 0.55 \\

\midrule

& \multicolumn{4}{c|}{\textbf{Qwen3-14B}} &
  \multicolumn{4}{c}{\textbf{Qwen3-14B}} \\
Baseline
& 0.23 & 3.00 & 7.6 & 0.72
& 0.34 & 7.39 & 1179 & 0.19 \\
TiO
& 0.29 & 4.12 & 7.0 & {0.90}
& 0.48 & 7.08 & 1332.4 & 0.26 \\
Dual-Goal
& 0.13 & 0.50 & 15.0 & 0.41
& 0.58 & 7.66 & 1191.7 & 0.40 \\
\midrule

& \multicolumn{4}{c|}{\textbf{Qwen3-32B}} &
  \multicolumn{4}{c}{\textbf{Qwen3-32B}} \\
Baseline
& 0.33 & 4.93 & 6.6 & 0.50
& 0.41 & 10.83 & 1029.8 & 0.14 \\
TiO
& 0.36 & 3.76 & {6.1} & 0.63
& {0.61} & 6.38 & 1450.0 & 0.54 \\
Dual-Goal
& 0.10 & {0.00} & 15.4 & 0.46
& 0.60 & 10.29 & 1472.3 & 0.77 \\
\midrule

& \multicolumn{4}{c|}{\textbf{LLaMA-3.3-70B-Ins}} &
  \multicolumn{4}{c}{\textbf{DeepSeek-R1-LLaMA-70B}} \\
Baseline
& 0.21 & 9.37 & 14.1 & 0.37
& 0.35 & 8.10 & 957.6 & 0.32 \\
TiO
& 0.38 & 10.03 & 14.7 & 0.50
& 0.46 & 10.07 & 1072.8 & 0.48 \\
Dual-Goal
& 0.20 & 12.87 & 23.7 & 0.53
& 0.50 & 8.39 & 1211.2 & 0.61 \\
\midrule

& \multicolumn{4}{c|}{\textbf{GPT-4o}} &
  \multicolumn{4}{c}{\textbf{QwQ-32B}} \\
Baseline
& 0.11 & 0.00 & 10.4 & 0.25
& 0.50 & 11.75 & 1757.5 & 0.36 \\
TiO
& 0.16 & 2.50 & 10.7 & 0.17
& 0.59 & 10.28 & 1464.5 & 0.37 \\
Dual-Goal
& 0.14 & 6.75 & 17.1 & 0.34
& 0.58 & 10.76 & 2277.2 & {0.81} \\

\midrule

& \multicolumn{4}{c|}{--} &
  \multicolumn{4}{c}{\textbf{Gemini-2.5-Pro}} \\
Baseline
& -- & -- & -- & --
& 0.73 & 11.14 & -- & 0.66 \\
TiO
& -- & -- & -- & --
& 0.78 & 8.52 & -- & 0.75 \\
Dual-Goal
& -- & -- & -- & --
& 0.78 & 10.10 & -- & 0.77 \\

\bottomrule
\end{tabular}
\end{table*}

\begin{wrapfigure}{r}{0.6\columnwidth}
\centering
\vspace{-0.3em}
\begin{tabular}{cc}
\includegraphics[width=0.28\columnwidth]{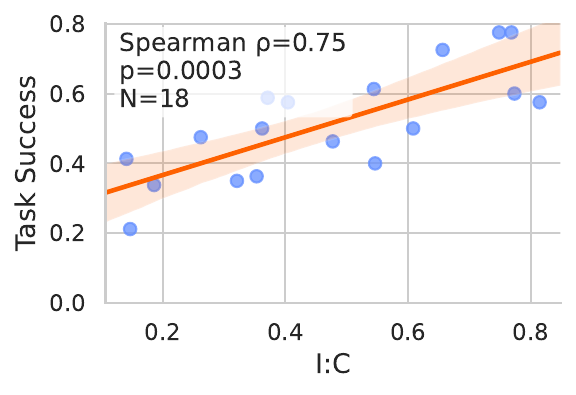} &
\includegraphics[width=0.28\columnwidth]{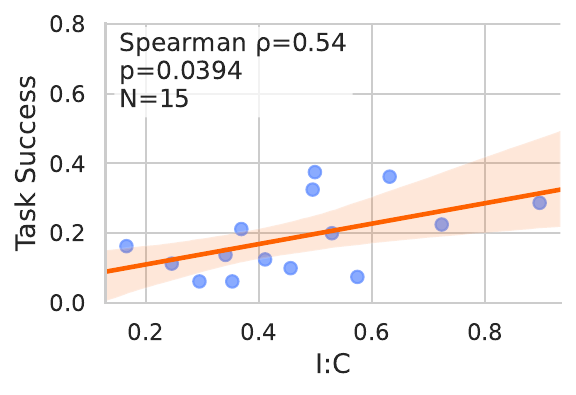}
\end{tabular}
\begin{tabular}{cc}
\small{(a) Thinking models} &
\small{(b) Non-thinking models}
\end{tabular}
\caption{\textbf{Confirmation bias correlates with task success.}
Higher I:C (more disconfirmatory testing) is associated with higher task success.
Each point represents a (model, variant) averaged over 80 episodes, and shaded regions show 95\% confidence intervals.
}
\label{fig:ic_correlations}
\vspace{-0.3em}
\end{wrapfigure}

Table~\ref{tab:replication} presents the experimental results in eleven LLMs. We see larger model variants are largely more successful within the same model family. In the first three row blocks, we can compare three models (Qwen3-8B, 14B, 32B) in thinking vs. non-thinking mode. We find that thinking mode largely improves task success rates over non-thinking mode. However, the number of turns to success is not always fewer with thinking models; potentially, they explore more to identify harder rules, which non-thinking models fail to guess.
We also find that switching non-thinking to thinking mode does not come with increase in I:C, indicating that gains from thinking mode are not driven by reduced confirmation bias.

Additional diagnostics, such as first attempt guess rate and results broken down by rule category, are in Appendix~\ref{appendix:full-results}. We verify that re-prompting frequency does not affect results in Appendix~\ref{appendix:retry}, 
that task success reflects genuine exploration quality rather than prior alignment in 
Appendix~\ref{appendix:controlled-prior}, and that the model's announcement
generally reflects its current hypothesis in Appendix~\ref{appendix:hypothesis-agreement}.

\noindent\textbf{Lower confirmation bias correlates with higher task success}
Figure~\ref{fig:ic_correlations} plots I:C ratios (x-axis) against task success rates (y-axis),
separately for thinking and non-thinking models.
We observe statistically significant positive correlations in both cases,
with a stronger trend for thinking models.
Overall, this indicates that reduced confirmation bias is associated with higher task success,
suggesting that mitigating confirmation bias in LLMs can improve its performance.

\begin{wrapfigure}{r}{0.6\columnwidth}
  \centering

  \begin{tabular}{cc}
    \includegraphics[width=0.28\columnwidth]{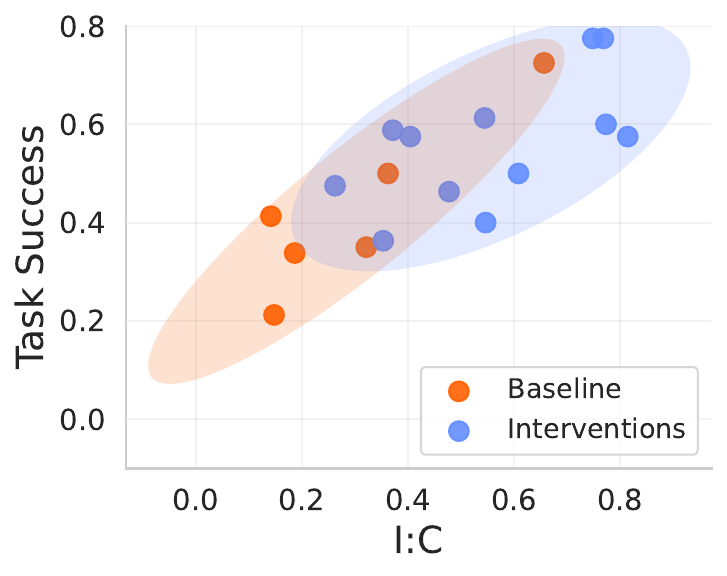} &
    \includegraphics[width=0.28\columnwidth]{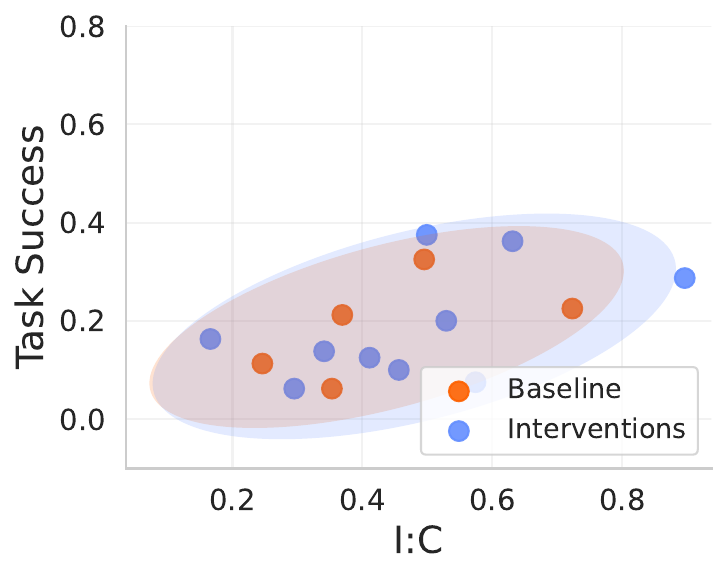}
  \end{tabular}


  \begin{tabular}{cc}
    \small (a) Thinking models &
    \small (b) Non-thinking models
  \end{tabular}

  \caption{\textbf{Interventions shift exploration in thinking but not non-thinking models.}
  Red denotes Baseline; blue denotes intervention runs (Dual-Goal / Think-in-Opposites).
  Ellipses show 95\% covariance regions.
  \emph{Each point represents a (model, variant) averaged over 80 episodes.}}
  \label{fig:intervention_shift}
  \vspace{-1.4em}
\end{wrapfigure}

\noindent\textbf{Impact of interventions}
Overall, the intervention strategies improves task success.
Dual-Goal improves task success in eight out of eleven model–inference scenarios, while Think-in-Opposites improves task success in all.
Both interventions tend to increase I:C ratios (18 out of 22 cases). Figure~\ref{fig:intervention_shift} visualizes these trends.

Statistical significance is assessed using paired permutation tests with $50{,}000$ permutations, with $p$-values adjusted via Benjamini--Hochberg correction; more details in Appendix~\ref{appendix:intervention-stats}. We found the improvements to be statistically significant for thinking models,
whereas for non-thinking models the results are mixed (no significant change for I:C, small improvement in accuracy for TiO and small drop in accuracy for Dual-Goal).

\section{Injecting Intervention-Guided Reasoning into LLMs}
\label{sec:distillation}
Having observed that the intervention prompt improves the LLM's task performance by reducing confirmation bias, we study how to internalize this without providing it at inference time as a prompt. We use a symbolic knowledge distillation approach~\citep{west2022symbolicknowledgedistillationgeneral}. 
We generate input-output pairs from settings that exhibit lower confirmation 
bias and higher task success rate: outputs produced under intervention prompts 
and outputs from larger teacher models. We perform 
supervised fine-tuning on that data.
We describe the set-up below. 

\noindent\textbf{Distillation setup}
To target exploration behavior rather than rule-guessing accuracy, we train on
\textbf{\texttt{Test} turns} produced by a teacher, including unsolved episodes and solved ones.
Each training example consists of the dialogue history up to a \texttt{Test} turn
and the teacher’s next-turn response, which includes reasoning tokens.

More formally, let $x_t$ denote the dialogue history up to the $t$-th \texttt{Test} turn.
Let $T_{\text{teacher}}(x_t)$ and $T_{\text{student}}(x_t)$ denote the prompts used for
the teacher and student, respectively.
The teacher model $p_{\theta_T}$ produces a next-turn response
$y^T = (y^T_1,\dots,y^T_{|y^T|})$ under $T_{\text{teacher}}(x_t)$.
For distillation, the student model $p_{\theta}$ is conditioned on the same dialogue
history formatted with $T_{\text{student}}(x_t)$ and trained to imitate the teacher
output token by token. 
Specifically, we optimize the log-likelihood:
\begin{equation*}
\mathcal{L}_{\textsc{sft}}(\theta)
=
- \mathbb{E}_{(x_t, y^T) \sim \mathcal{D}}
\sum_{i=1}^{|y^T|}
\log p_{\theta}\!\left(y^T_i \mid T_{\text{student}}(x_t), y^T_{<i}\right)
\end{equation*}

We construct the distillation dataset using the train/validation/test split defined in Section~\ref{sec:framework}. Teacher-generated episodes from the
training split yield $4 \times 4 \times 100 = 1600$ episodes, each with 45
\texttt{Test} turns, resulting in $72{,}000$ training examples.
Validation uses teacher-generated episodes from the validation split,
yielding $16 \times 45 = 720$ validation examples. Training details are in Appendix~\ref{appendix:distillation-details}.

\noindent\textbf{Teacher--student Set-ups}
We focus on Think-in-Opposites for distillation, as Dual-Goal alters the output
format (DAX/MED rules).
We study the following teacher--student configurations:
\begin{itemize}[noitemsep,leftmargin=10px]
\item \textbf{Intervention:} a model distilling from its own outputs generated with Think-in-Opposites prompts.
\item \textbf{Cross-scale:} a student
distilling from a larger teacher model, with the same baseline prompt.
\item \textbf{Cross-scale and Intervention:} a smaller student
distilled from a larger model using Think-in-Opposites prompts.
\end{itemize}

\begin{wrapfigure}{r}{0.52\textwidth}
\vspace{-3em}
\centering
\includegraphics[width=0.98\linewidth]{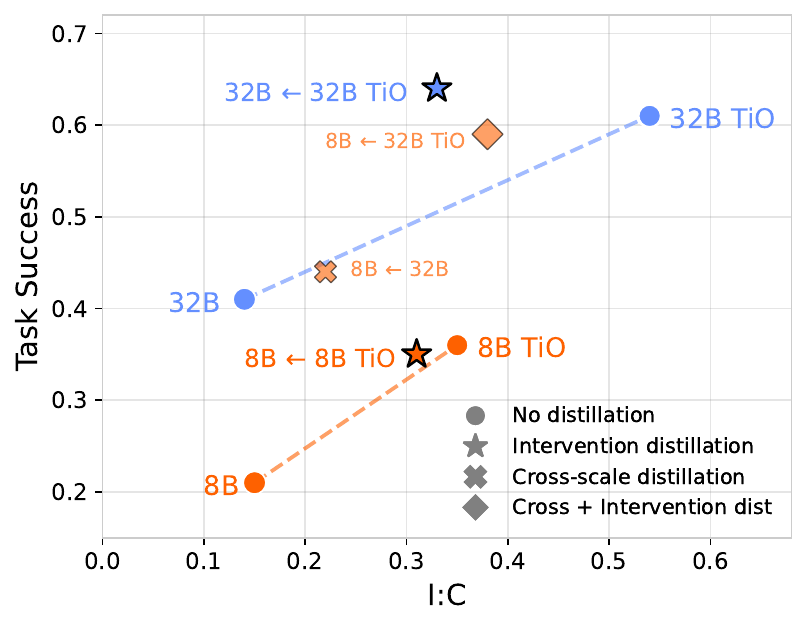}
\caption{
Distillation results for Qwen3 models on the Wason task. Distillation is denoted as \emph{Student $\leftarrow$ Teacher}. Red: 8B student; Blue: 32B student.
}
\label{fig:wason-distill}
\vspace{-1em}
\end{wrapfigure}






\paragraph{Results}
Figure~\ref{fig:wason-distill} shows the distillation performance (full results in Appendix~\ref{appendix:wason-distillation-results}). 
We observe that intervention distillation improves task success within the same base model, raising it to a comparable performance to its teacher setting.

Cross-scale distillation improves both task performance and I:C ratio. Unsurprisingly, the largest gain was achieved from cross-scale, intervention distillation, almost tripling the task success rates and more than doubling the I:C ratio. Overall, we report evidence of within-task generalization across different rules. 

We further perform a sanity-check whether this distillation hurts other abilities of LLMs, by evaluating it on a mathematical reasoning benchmark (GSM8K~\cite{cobbe2021training}). We observe little (at most 2\%) or no degradation (results in Appendix~\ref{appendix:gsm8k}). We next investigate whether these distilled models generalize to a new task where LLMs suffer from confirmation bias.

\section{Generalization to a New Task Domain: Blicket Test}


We study {the Blicket Test} environment from \citet{gxchen2025language}, another task that requires hypothesis testing. Designed by developmental psychologists to test children’s ability for causal reasoning \citep{gopnik2000detecting}, this setting has recently been adapted into text-based games to evaluate LLMs \citep{piriyakulkij2024doing,gxchen2025language}.

\noindent\textbf{Task} It involves $N$ objects and a ``Blicket detector'' machine. A subset of objects (the ``Blickets'') activates the machine according to an unobserved rule: under the OR rule, the machine turns on when any Blicket object is placed on it, while under the AND rule, it turns on only when \textit{all} Blicket objects are placed on it. This tests an agent's ability to perform exploration and causal discovery.


We adapt the Blicket test to our interactive hypothesis-testing framework (see Appendix~\ref{appendix:blicket-adaptation} for comparison with the original experiment).
Each \textbf{episode} starts with an initial object placement and the machine being ON or OFF, followed by 45 alternating \texttt{Guess} and \texttt{Test} turns, mirroring our Wason setup.
On \texttt{Test} turns, the model selects objects to place on the machine and receives feedback whether it turns ``ON'' or ``OFF''.
On \texttt{Guess} turns, the model guesses the blickets and hidden rule; no feedback is provided.
A correct guess must identify both the full set of blickets and the correct rule. We evaluate a Baseline prompt and an adapted Think-in-Opposites (TiO) prompt (provided in Appendix~\ref{appendix:blicket-prompt} with example interactions). We do not evaluate Dual-Goal because it requires an opposite hypothesis, while a Blicket hypothesis (set of blickets + rule) has no well-defined opposite.

\noindent\textbf{Data}
We vary the task along three dimensions: (1) number of objects (4 or 8), (2) number of blickets (2 or 3), and (3) hidden rule (AND, OR, or XOR). 
For each configuration, we generate 16 episodes by varying the blicket assignments and initial object placements.
This yields $2 \times 2 \times 3 \times 16 = 192$ evaluation episodes in total.
Dataset construction details and statistics are provided in Appendix~\ref{appendix:blicket-dataset}.

\begin{table*}[t]
\centering

\begin{minipage}[t]{0.60\textwidth}
\vspace{0pt}

\caption{\textbf{LLM Task performance in the Blicket Task.} }
\label{tab:blicket_main}

\centering
\small
\setlength{\tabcolsep}{1.5pt}
\renewcommand{\arraystretch}{0.9}

\begin{tabular}{l| c c c | c c c}
\toprule
& \multicolumn{3}{c|}{\textbf{Non-Thinking}} &
  \multicolumn{3}{c}{\textbf{Thinking}} \\
\cmidrule(lr){2-4} \cmidrule(lr){5-7}
& \textbf{Task Suc.} &
  \textbf{\# Turns} &
  \textbf{I:C} &
  \textbf{Task Suc.} &
  \textbf{\# Turns} &
  \textbf{I:C} \\
& \textbf{\% $\uparrow$} &
  \textbf{until Suc. $\downarrow$} &
  \textbf{$\uparrow$} &
  \textbf{\% $\uparrow$} &
  \textbf{until Suc. $\downarrow$} &
  \textbf{$\uparrow$} \\
\midrule

& \multicolumn{3}{c|}{\textbf{Qwen3-8B}} &
  \multicolumn{3}{c}{\textbf{Qwen3-8B}} \\
Baseline & 0.11 & 10.24 & 0.36 & 0.32 & 10.26 & 0.44 \\
TiO      & 0.19 & 12.14 & 0.25 & 0.40 & 10.73 & 0.57 \\
\midrule

& \multicolumn{3}{c|}{\textbf{Qwen3-14B}} &
  \multicolumn{3}{c}{\textbf{Qwen3-14B}} \\
Baseline & 0.25 & 10.38 & 0.55 & 0.50 &  9.48 & 0.34 \\
TiO      & 0.20 & 10.73 & 0.74 & 0.61 & 12.33 & 0.48 \\
\midrule

& \multicolumn{3}{c|}{\textbf{Qwen3-32B}} &
  \multicolumn{3}{c}{\textbf{Qwen3-32B}} \\
Baseline & 0.64 & 14.00 & 1.00 & 0.57 & 12.18 & 0.54 \\
TiO      & 0.56 & 13.89 & 1.33 & \textbf{0.86} & 12.98 & \textbf{1.31} \\
\midrule

& \multicolumn{3}{c|}{\textbf{LLaMA-3.3-70B-Ins}} &
  \multicolumn{3}{c}{\textbf{DeepSeek-R1-LLaMA-70B}} \\
Baseline & 0.78 & 12.91 & 0.82 & 0.41 & 12.93 & 0.43 \\
TiO      & 0.82 & 13.50 & 0.98 & 0.43 & 17.41 & 0.57 \\
\midrule

& \multicolumn{3}{c|}{\textbf{--}} &
  \multicolumn{3}{c}{\textbf{QwQ-32B}} \\
Baseline & -- & -- & -- & 0.80 & 12.27 & 0.89 \\
TiO      & -- & -- & -- & 0.82 & 14.68 & 1.30 \\
\bottomrule
\end{tabular}

\end{minipage}
\hfill
\begin{minipage}[t]{0.36\textwidth}
\vspace{0pt}
\centering

\includegraphics[width=0.9\linewidth]{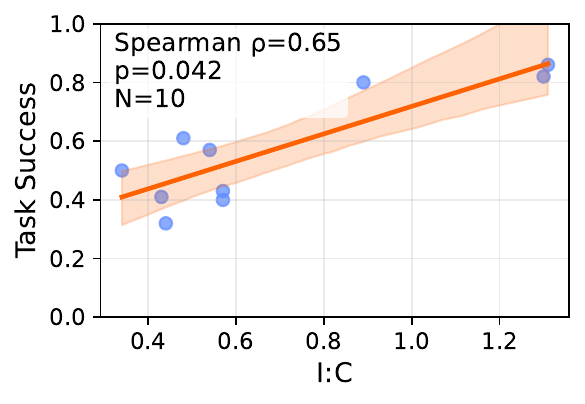}

\vspace{-0.6em}
{\footnotesize (a) Thinking models}


\includegraphics[width=0.9\linewidth]{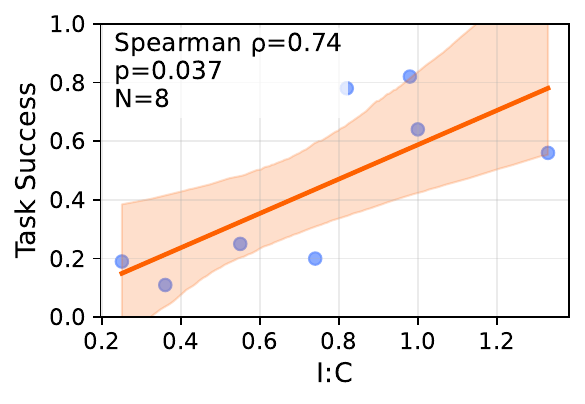}

\vspace{-0.6em}
{\footnotesize (b) Non-thinking models}

\captionof{figure}{\textbf{Correlation between the extent of confirmation bias and task success.}}
\label{fig:ic_correlations_blickets}

\end{minipage}

\end{table*}




\noindent\textbf{Evaluation}
Following the evaluation setup in the Wason task (described in Sec~\ref{subsec:eval_metrics}), we report: (1) task success, (2) \# turns until success, and (3) I:C ratio. 
We keep the same hyperparameters as the Wason task, and modify the LLM-as-a-Judge prompt (provided in Appendix~\ref{appendix:blicket-judge-prompts} with examples).

\noindent\textbf{Does confirmation bias correlate negatively with task success in this domain?}
We first evaluate whether the relationship between confirmation bias and task performance observed in the Wason task generalizes to the Blicket test.
Table~\ref{tab:blicket_main} reports overall performance across models, and Figure~\ref{fig:ic_correlations_blickets} plots I:C ratios against task success. We observe significant positive correlations between I:C and task success for both thinking and non-thinking models, indicating that models that show reduced confirmation bias tend to solve the task more successfully. Finally, we find that lower confirmation bias is associated with better task
performance overall. This trend is visible for both thinking and non-thinking
models, and is slightly stronger among non-thinking models in terms of rank
correlation.

\noindent\textbf{Does the intervention strategy to reduce confirmation bias enhance task performance in this domain?}
We next evaluate whether Think-in-Opposites (TiO) improves performance in the Blicket task.
Across thinking models, TiO increases both task success and I:C on average, and these improvements are statistically significant under paired permutation tests (Appendix~\ref{appendix:intervention-stats}). 
For non-thinking models, the effects are substantially smaller.

\section{Generalization of Distilled Model}

We evaluate four models: base Qwen3-8B and Qwen3-32B, and these models further fine-tuned on outputs generated with Think-in-Opposites prompts (Qwen3-8B + 8B TiO and Qwen3-32B + 32B TiO) from Section \ref{sec:distillation}.

\subsection{Generalization to Blicket Test}


\begin{wrapfigure}{r}{0.46\textwidth}
\vspace{-1.3em}
\centering
\includegraphics[width=0.8\linewidth]{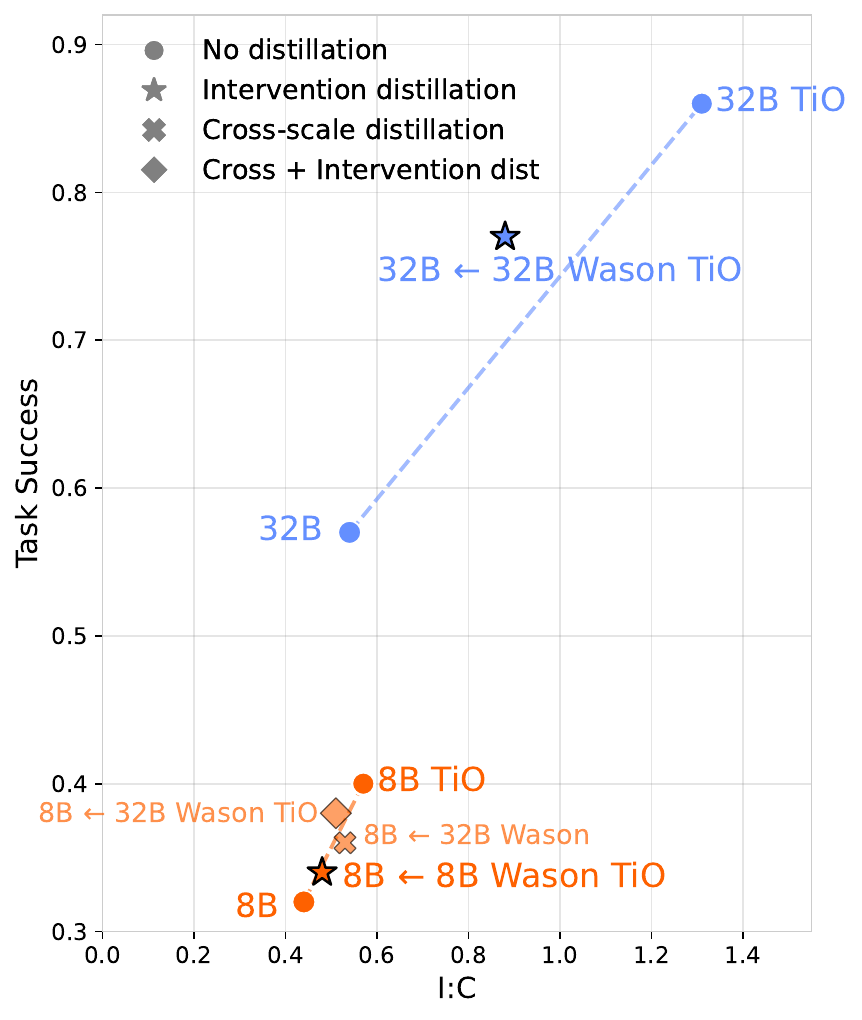}
\caption{
Blicket task performance for \textbf{Qwen3 models} distilled on the Wason task. Distillation is denoted as \emph{Student $\leftarrow$ Teacher}.
}
\label{fig:blicket-distill}
\vspace{-2em}
\end{wrapfigure} 
Would training to decrease confirmation bias in the number game generalize to the new object-based domain? Figure~\ref{fig:blicket-distill} shows Blicket performance for models distilled on the Wason task, with base model results shown for comparison (full results in Appendix~\ref{appendix:blicket-transfer-results}). Same-scale intervention distillation (8B TiO $\rightarrow$ 8B student; 32B TiO $\rightarrow$ 32B student) yields  improvements over the corresponding base models, with statistical test results provided in Appendix~\ref{appendix:distillation-stats}. However, performance remains below applying TiO prompting in the Blicket task. This suggests that falsification-oriented exploration learned in the Wason task transfers to a new domain and leads to measurable gains, although it does not fully match the improvements from task-specific intervention prompts.

\subsection{Generalization to Mathematical Self Correction}
\label{sec:math-generalization}

We next ask whether this behavior generalizes beyond rule discovery. In mathematical reasoning, a model given another attempt after an incorrect answer may repeat its original reasoning rather than explore an alternative solution path, arriving at the same incorrect answer. This behavior is analogous to confirmation bias, where an agent seeks evidence supporting an incorrect hypothesis rather than testing alternatives. We therefore evaluate whether models distilled on the Wason task are better able to revise their solutions after being told their previous answer was incorrect.

\noindent\textbf{Setting}
We evaluate on AIME 2025~\citep{opencompass2025aime} and Beyond AIME~\citep{bytedance_seed_2025_beyondaime}, two mathematical reasoning benchmarks. Their problems are sufficiently challenging that models often fail on the first attempt, creating opportunities for iterative revision. For each problem, the model first attempts a solution. If its answer is incorrect, we retain the conversation history, inform the model that its previous answer was incorrect, and ask it to review its previous reasoning and try again for up to five times. Full prompt and evaluation details are provided in Appendix~\ref{appendix:math-reasoning}.



\begin{figure*}[h]
\centering
\includegraphics[width=0.8\linewidth]{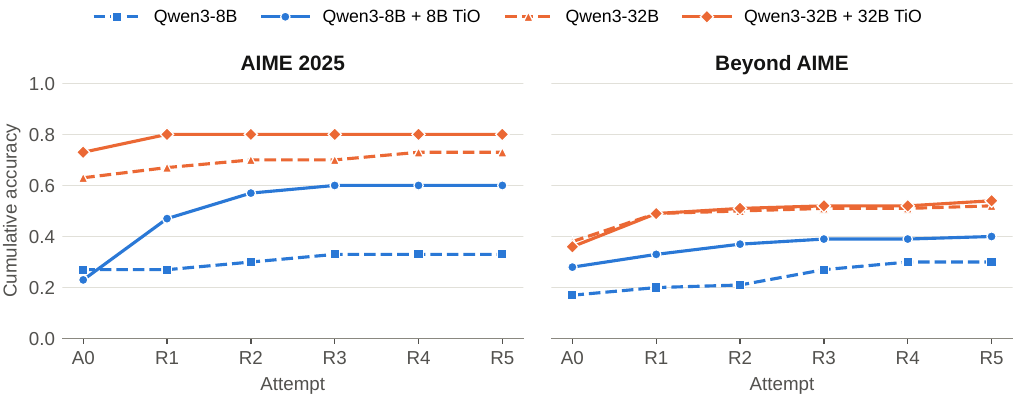}
\caption{\textbf{Mathematical reasoning accuracy across retries.}
Accuracy is cumulative: at each retry, a problem is counted as solved if the model has produced the correct answer in any attempt up to that point. A0 denotes the initial attempt (Attempt 0), and R1--R5 denote Retries 1--5. Distilled models (Qwen3-8B + 8B TiO, Qwen3-32B + 32B TiO) were trained on the Wason task.}
\label{tab:math-retries}
\end{figure*}

\paragraph{Results} Figure~\ref{tab:math-retries} shows that, for the 8B model, the distilled version improves final accuracy over the base model on both benchmarks (0.60 vs. 0.33 on AIME 2025; 0.40 vs. 0.30 on Beyond AIME). Using a two-sided exact McNemar test on final solved/unsolved outcomes, this improvement is statistically significant on both ($p=0.0386$ and $p=0.0414$, respectively). For the 32B model, there are smaller gains (0.80 vs. 0.73 on AIME 2025; 0.54 vs. 0.52 on Beyond AIME), but these are not statistically significant ($p=0.6875$ and $p=0.7539$, respectively), likely reflecting less room for self-correction at the higher baseline accuracy. These results suggest that the debiasing signal learned on the Wason task carries over to more effective self-correction in math.



\section{Related Work}

Adapting tools from cognitive science for identifying cognitive biases in humans can be an effective approach for identifying similar issues in LLMs. LLMs have been shown to reflect a wide range of biases found in humans, such as content effects \citep{dasgupta2022language}, threshold priming \citep{10.1145/3673791.3698420}, overthinking \citep{liu2025mind}, overestimating rationality \citep{liu2025large}, forming implicit associations \citep{bai2025explicitly}, linguistic illusions \citep{marjieh2024rational}, skewed representations of numbers \citep{marjieh2025number}, conformity under social pressure \citep{zhu2025conformity}, and causal reasoning \citep{gxchen2025language}.

Confirmation bias is one of well-established cognitive biases in humans, from seminal Wason's 2-4-6 rule-discovery task \citep{wason1960failure}. 
In the subsequent decades, cognitive scientists have studied identifying and mitigating confirmation bias in humans \citep{gale2006,branchini2023}.
Recent work has explored confirmation bias in LLMs as part of broader bias evaluations~\citep{malberg2025comprehensive}, in chain-of-thought reasoning~\citep{wan2025unveiling}, in mitigating hallucinations~\citep{ok2025synthetic}, in LLM-assisted security code review~\citep{mitropoulos2026measuring}, and in relation to framing and position biases~\citep{ermakova2026confirmation}. 
However, these studies primarily examine confirmation bias at the level of stance or answer generation—for example, whether a model favors supporting over contradicting arguments once a claim or hypothesis has already been presented. 
Such settings measure confirmation bias in evaluative reasoning.

In contrast, we study confirmation bias during the {exploration process}. 
Our setting requires models to iteratively generate hypotheses, select which evidence to test, and revise beliefs based on environment feedback. 
Here, confirmation bias manifests not in how evidence is interpreted, but in how evidence is {selected}. 
This distinction is critical for agentic systems that must autonomously explore hypothesis spaces, such as in AI-driven scientific discovery~\citep{romera2024mathematical,yamada2025aiscientist,novikov2025alphaevolve,castro2025discovering,aygun2025aiscientist}.

Related work has studied hypothesis exploration through experiment design. \citet{piriyakulkij2024doing} use expected information gain to identify informative tests given a distribution over hypotheses, requiring an explicit hypothesis space and posterior. 



\section{Conclusion}
\label{sec:conclusion}
We adapt the experimental settings from cognitive psychology to study confirmation bias in LLMs 
during interactive hypothesis exploration. Introducing a controlled rule-discovery 
framework and the I:C metric, we show that all 11 LLMs we evaluate favor 
confirmatory tests, and that reducing this bias leads to gains 
in rule discovery. Distilled models further generalize this falsification-oriented 
behavior to a new rule-discovery domain without additional prompting and beyond rule discovery 
to mathematical self-correction, suggesting that debiasing can be internalized at the parameter level.

\paragraph{Limitations and Future Work}
Our framework shows how to measure and mitigate confirmation bias with a focus on rule-discovery settings. 

While we find that the benefits of distillation extend beyond rule discovery to mathematical self-correction, an important direction for future work is evaluating whether similar improvements generalize to other real-world settings, such as scientific discovery, question answering, and agentic tasks involving planning, tool use, and debugging.

\section*{Acknowledgements and Disclosures}
 We thank Nishant Balepur, Kristina Fujimoto, and John (Yueh-Han) Chen for their feedback. 
This work was supported in part through
the NYU IT High Performance Computing resources, services, and staff expertise. The work is partially funded by NSF CAREER award 2443271. We also thank Google for providing Gemini API credits. The authors declare no competing interests. 

\bibliography{neurips_2026.bib}
\bibliographystyle{plainnat}

\appendix
\section{Dataset Construction Details}
\label{appendix:data}

\subsection{Prompt for Rule Generation}
\label{appendix:ruleprompt}

We generate rule groups using a large language model prompted to propose relational hypotheses over integer triples $(a,b,c)\in[-99,100]^3$.
The model is instructed to generate \textbf{ten rule groups} in a single response, where each group contains \textbf{four distinct rules}.
All four rules within a group are required to share a non-empty feasible intersection (i.e., at least one triple satisfies all four rules).

\begin{PromptBlock}
    
You are generating relational rules over integer triples (a, b, c) for a reasoning task.\\

Each rule should describe a simple, interpretable relation or shared property among the three integers.
Avoid unary conditions that describe only one of the numbers (e.g., ``b is even'' or ``a is positive'').
Involve all three variables jointly — for example, patterns or equations that connect a, b, and c together — rather than isolated or purely pairwise comparisons.\\

Rules may belong to any of the following families:\\
- Ordering (ascending, descending, non-decreasing, non-monotone, etc.)\\
- Arithmetic structure (equal differences, geometric relations, increasing or decreasing gaps, sum relationships, etc.)\\
- Parity or divisibility (even/odd, divisibility chains, modular patterns, etc.)\\
- Sign (all positive, all negative, mixed signs)\\
- Number properties (multiples, primes, squares, or related characteristics)\\
- Comparative balance (averages, sums, or ratios involving all three numbers)\\
- Symmetry or modular relations (mirror patterns, alternating parities, repeating residues)\\

Do not use rules that depend on numeric ranges (e.g., ``between 10 and 50'' or ``less than 100'').\\

Each rule should naturally vary in both:\\
- Breadth: the proportion of triples in $[-99, 100]^3$ that satisfy it.\\
- Rule-guess difficulty: how easily a human or model might infer it.\\

Now construct a table of rule groups as follows:\\
- Each group contains four distinct rules.\\
- All four rules in a group must have a non-empty feasible intersection.\\
- All rules across all groups must be distinct (no repetitions or paraphrases).\\
- Rules in a group may belong to different families.\\
- Avoid groups where all rules are extremely narrow; ensure at least moderate feasible overlap.\\

Generate 10 groups of 4 rules each. Output only the Markdown table, no LaTeX or math mode.\\
\end{PromptBlock}

Rules depending on explicit numeric ranges are disallowed, as they introduce arbitrary constraints rather than abstract relational structure.

\subsection{Post-filtering and Revision}
\label{appendix:filtering}

After automatic generation, we apply structured post-filtering and revision.
A rule group is rejected and repaired if it violates any of the following criteria:
\begin{enumerate}
    \item \textbf{Insufficient feasible intersection:} the feasible intersection set contains fewer than five valid triples.
    \item \textbf{Unary dependence:} the rule constrains only a single variable (e.g., ``$c$ is prime'').
    Conjunctive properties applied uniformly to all three variables (e.g., ``all odd'') are retained.
    \item \textbf{Excessive complexity:} the rule requires multi-step arithmetic dependencies unlikely to arise during hypothesis testing.
\end{enumerate}

When a group fails any criterion, we re-prompt the language model to replace only the offending rule(s) while keeping the remaining rules fixed.
This process is repeated until all constraints are satisfied.
Approximately one-third of initially generated groups required at least one revision.

After filtering, we randomly shuffle the order of the rule groups using a fixed seed (1337), without modifying the rules within each group.

\subsection{Human Rule Injection}
\label{appendix:humanrules}

To incorporate human-like hypotheses, we inject one human-derived rule into each rule group.
Human rules are drawn from prior studies on number generalization, including the Large Dataset of Generalization Patterns in the Number Game~\citep{DVN/A8ZWLF_2018}, which builds on Tenenbaum’s Number Game paradigm~\citep{Tenenbaum2000RulesSimilarity}.

For each unique triple in the source dataset, multiple human participants provided pattern descriptions.
We select the most frequent description per triple and remove hypotheses referring to explicit numeric ranges.
From the remaining pool, we curate ten interpretable human hypotheses:
\emph{even}, \emph{odd}, \emph{prime}, \emph{cube}, \emph{ends with 1}, \emph{ends with 6}, \emph{ends with 9}, \emph{divisible by 3}, \emph{divisible by 4}, and \emph{divisible by 5}.

For each rule group, we replace one LLM-generated rule with a selected human-derived rule.
The replacement is accepted only if the resulting group maintains a non-empty feasible intersection.
Human-derived rules are indicated explicitly in Table~\ref{tab:groupings}.

\subsection{Feasible Intersection and Triple Sampling}
\label{appendix:sampling}

For each rule group $g$ with rules $\{r_{g,i}\}_{i=1}^4$, we define the feasible set
$S_g = \{(a,b,c)\in[-99,100]^3 \mid r_{g,i}(a,b,c)=1 \ \forall i\}$.

We compute $|S_g|$ by enumerating all triples in $[-99,100]^3$ and evaluating all four rule predicates.
The feasible set size for each group is reported in Table~\ref{tab:groupings}.

We uniformly sample triples from $S_g$ using a fixed random seed (1337).
Each sampled triple serves as the initial triple for four reasoning episodes, one per rule in the corresponding group treated as the target rule.

\subsection{Dataset Splits and Scenario Counts}
\label{appendix:splits}

We divide the ten rule groups into \textbf{4 Train}, \textbf{2 Validation}, and \textbf{4 Test} groups.
Training uses 100 triples per group, yielding $4 \times 100 \times 4 = 1600$ training scenarios.
Validation uses 2 triples per group, yielding $2 \times 2 \times 4 = 16$ validation scenarios.
Testing uses 5 triples per group; since each triple is evaluated against all four target rules, this yields $4 \times 5 \times 4 = 80$ test scenarios.

Rule definitions, executable predicates, rule groupings, feasible intersection sizes, and sampled test instances are provided in Tables~\ref{tab:rules-train}--\ref{tab:instances-test}.

\begin{table}[H]
\centering
\small
\caption{Rules (Train).}
\label{tab:rules-train}
\begin{tabularx}{\textwidth}{>{\raggedright\arraybackslash}p{0.33\textwidth} X}
\toprule
\textbf{Rule} & \textbf{Predicate (over a,b,c)} \\
\midrule
All even & \texttt{a\%2==0 and b\%2==0 and c\%2==0} \\
Each divides next & \texttt{a!=0 and (b\%a==0) and b!=0 and (c\%b==0)} \\
Exactly two equal & \texttt{len(\{a,b,c\})==2} \\
At least one even & \texttt{(a\%2==0) or (b\%2==0) or (c\%2==0)} \\
All end with 6 & \texttt{abs(a)\%10==6 and abs(b)\%10==6 and abs(c)\%10==6} \\
Increasing differences & \texttt{0 \textless{} (b-a) \textless{} (c-b)} \\
a is min & \texttt{a<=b and a<=c} \\
All distinct & \texttt{len(\{a,b,c\})==3} \\
All divisible by 5 & \texttt{a\%5==0 and b\%5==0 and c\%5==0} \\
a is max & \texttt{a>=b and a>=c} \\
Non-monotone (middle between ends) & \texttt{(a-b)*(b-c)\textless{}0} \\
At least two multiples of 5 & \texttt{((a\%5==0)+(b\%5==0)+(c\%5==0))>=2} \\
All divisible by 3 & \texttt{a\%3==0 and b\%3==0 and c\%3==0} \\
Alternating parity (ends same) & \texttt{(a\%2)==(c\%2) and (b\%2)!=(a\%2)} \\
Ascending & \texttt{a\textless{}b\textless{}c} \\
Non-decreasing & \texttt{a<=b<=c} \\
\bottomrule
\end{tabularx}
\end{table}
\begin{table}[H]
\centering
\small
\caption{Rules (Validation).}
\label{tab:rules-val}
\begin{tabularx}{\textwidth}{>{\raggedright\arraybackslash}p{0.33\textwidth} X}
\toprule
\textbf{Rule} & \textbf{Predicate (over a,b,c)} \\
\midrule
All end with 9 & \texttt{abs(a)\%10==9 and abs(b)\%10==9 and abs(c)\%10==9} \\
Non-decreasing differences & \texttt{(b-a) <= (c-b)} \\
c is max & \texttt{c>=a and c>=b} \\
Arithmetic progression (AP) & \texttt{(b-a)==(c-b)} \\
All divisible by 7 & \texttt{a\%7==0 and b\%7==0 and c\%7==0} \\
Exactly two even & \texttt{((a\%2==0)+(b\%2==0)+(c\%2==0))==2} \\
At least one multiple of 4 & \texttt{(a\%4==0) or (b\%4==0) or (c\%4==0)} \\
At least two distinct & \texttt{len(\{a,b,c\})>=2} \\
\bottomrule
\end{tabularx}
\end{table}

\begin{table}[H]
\centering
\small
\caption{Rules (Test).}
\label{tab:rules-test}
\begin{tabular}{lp{0.62\linewidth}}
\toprule
\textbf{Rule} & \textbf{Predicate (over a,b,c)} \\
\midrule
All end with 1 & \texttt{abs(a)\%10==1 and abs(b)\%10==1 and abs(c)\%10==1} \\
c is min & \texttt{c<=a and c<=b} \\
All negative & \texttt{a<0 and b<0 and c<0} \\
Descending & \texttt{a>b>c} \\
All odd & \texttt{a\%2==1 and b\%2==1 and c\%2==1} \\
b is (strict) max & \texttt{b>a and b>c} \\
Mixed signs & \texttt{((a<0)+(b<0)+(c<0)) in \{1,2\}} \\
At least one multiple of 3 & \texttt{(a\%3==0) or (b\%3==0) or (c\%3==0)} \\
All prime numbers & \texttt{is\_prime(a) and is\_prime(b) and is\_prime(c)} \\
Non-increasing & \texttt{a>=b>=c} \\
All positive & \texttt{a>0 and b>0 and c>0} \\
Contains a prime & \texttt{is\_prime(a) or is\_prime(b) or is\_prime(c)} \\
All cube numbers & \texttt{is\_cube(a) and is\_cube(b) and is\_cube(c)} \\
Exactly two odd & \texttt{((a\%2==1)+(b\%2==1)+(c\%2==1))==2} \\
Decreasing gaps & \texttt{(b-a)>(c-b)} \\
At least one odd & \texttt{(a\%2==1) or (b\%2==1) or (c\%2==1)} \\
\bottomrule
\end{tabular}
\end{table}

\begin{table}[H]
\centering
\small
\setlength{\tabcolsep}{3pt}
\renewcommand{\arraystretch}{1.15}
\caption{Rule groups and their feasible intersections. The rightmost column reports the number of triples $|S_g|$ in $[-99,100]^3$ that satisfy all four rules in the group.}
\label{tab:groupings}
\begin{tabularx}{\textwidth}{c c
>{\raggedright\arraybackslash}X
>{\raggedright\arraybackslash}X
>{\raggedright\arraybackslash}X
>{\raggedright\arraybackslash}X
r}
\toprule
\textbf{ID} & \textbf{Split} & \textbf{Rule 1 (Human)} & \textbf{Rule 2} & \textbf{Rule 3} & \textbf{Rule 4} & \textbf{Feasible $|S_g|$} \\
\midrule
1  & Train      & All even           & Each divides next                  & Exactly two equal                   & At least one even            & 1{,}629 \\
2  & Train      & All end with 6     & Increasing differences             & a is min                            & All distinct                 & 5{,}394 \\
3  & Train      & All divisible by 5 & a is max                           & Non-monotone (middle between ends)  & At least two multiples of 5  & 128{,}080 \\
4  & Train      & All divisible by 3 & Alternating parity (ends same)     & Ascending                           & Non-decreasing               & 194 \\
\midrule
5  & Validation & All end with 9     & Non-decreasing differences         & c is max                            & Arithmetic progression (AP)   & 2{,}550 \\
6  & Validation & All divisible by 7 & Exactly two even                   & At least one multiple of 4          & At least two distinct         & 72 \\
\midrule
7  & Test       & All end with 1     & c is min                           & All negative                        & Descending                   & 1{,}225 \\
8  & Test       & All odd            & b is (strict) max                  & Mixed signs                         & At least one multiple of 3    & 176{,}715 \\
9  & Test       & All prime numbers  & Non-increasing                     & All positive                        & Contains a prime             & 3{,}071 \\
10 & Test       & All cube numbers   & Exactly two odd                    & Decreasing gaps                     & At least one odd             & 205 \\
\bottomrule
\end{tabularx}
\end{table}

\begin{table}[H]
\centering
\small
\caption{Sampled Test instances (groups 7--10). Each triple is used as a starting point for four episodes, one per rule in the corresponding group.}
\label{tab:instances-test}
\begin{tabular}{ccc}
\toprule
\textbf{Split} & \textbf{Group ID} & \textbf{Triple} \\
\midrule
Test & 7 & $(-1,\,-81,\,-91)$ \\
Test & 7 & $(-61,\,-71,\,-91)$ \\
Test & 7 & $(-1,\,-21,\,-81)$ \\
Test & 7 & $(-21,\,-41,\,-91)$ \\
Test & 7 & $(-11,\,-31,\,-41)$ \\
\midrule
Test & 8 & $(-77,\,75,\,-47)$ \\
Test & 8 & $(-9,\,55,\,-71)$ \\
Test & 8 & $(-51,\,37,\,-87)$ \\
Test & 8 & $(-67,\,75,\,25)$ \\
Test & 8 & $(-39,\,81,\,3)$ \\
\midrule
Test & 9 & $(97,\,67,\,37)$ \\
Test & 9 & $(97,\,73,\,67)$ \\
Test & 9 & $(89,\,67,\,13)$ \\
Test & 9 & $(79,\,43,\,31)$ \\
Test & 9 & $(89,\,67,\,2)$ \\
\midrule
Test & 10 & $(-27,\,8,\,27)$ \\
Test & 10 & $(8,\,27,\,-27)$ \\
Test & 10 & $(-27,\,0,\,1)$ \\
Test & 10 & $(-1,\,64,\,-1)$ \\
Test & 10 & $(-64,\,1,\,-1)$ \\
\bottomrule
\end{tabular}
\end{table}

\subsection{Divergence from Original Wason Study}\label{appendix:divergencewason}

The task setting diverges from the original Wason \emph{2-4-6} study in three ways.
First, human participants flexibly choose between testing and guessing, whereas LLMs alternate between \texttt{Guess} and \texttt{Test} turns. This design is to make it easier to evaluate whether the LLM proposed a triple that is confirmatory or not to the current hypothesis.
Second, feedback is provided only on \texttt{Test} turns, while \texttt{Guess} turns receive no immediate feedback.
Third, human sessions terminate upon a correct rule announcement, the participant giving up, or a time limit, whereas LLM episodes run for a fixed number of turns independent of guess correctness.

\section{Task Instruction Prompts and Examples}
\label{appendix:prompts}

This appendix lists the task instructions provided to models in each interaction setting, with example continuations.

\subsection{Baseline Prompt}
\label{appendix:baseline-prompt}

\begin{PromptBlock}
You will be given three numbers which conform to a simple rule that I have in mind.
This rule is concerned with the common attributes or relations between any three numbers,
and not with their absolute magnitude – for example, it is not a rule like ``all numbers above (or below) 50.''\\

Your goal is to discover this hidden rule by proposing sets of three numbers.
You will then be told if the triple conforms with the hidden rule or not.\\

At each turn, you may perform one of two actions:

- Test: propose a new set of three integers, denoted [a,b,c]. You will then be told whether that triple conforms with the hidden rule (``YES'' or ``NO'').\\
- Announce: based on all previous tests, state what you believe the hidden rule is. You will not receive any feedback for announcements.\\

Each instruction you receive will specify what to do.\\

Format (must follow exactly):

- If the instruction is Turn - Announce, output exactly one line:
  Announce: \textless one short sentence naming the rule\textgreater

- If the instruction is Turn - Test, output exactly one line:
  Check: [a,b,c]\\

A triple that conforms with the hidden rule is: [12, -36, -36].\\
Let's begin.\\
Turn - Announce.
\end{PromptBlock}

\paragraph{Example interaction (hidden rule: all numbers are even).}
\emph{The following transcript continues directly from the final ``Turn--Announce'' in the initial prompt above.}

\begin{verbatim}
[model] Announce: The second and third numner are equal.

[environment] Turn - Test

[model] Check: [2, 4, 4]

[environment] YES. Turn - Announce

[model] Announce: The second and third numner are equal.

[environment] Turn - Test

[model] Check: [4, 6, 6]

[environment] YES. Turn - Announce

[model] Announce: The second and third numner are equal.
\end{verbatim}

\subsection{Dual-Goal Prompt}
\label{appendix:dualgoal-prompt}

\begin{PromptBlock}
You will be given three numbers which conform to a simple rule that I have in mind.
Triples that fit this rule are called DAX, and triples that do not fit this rule are called MED.\\

These rules are concerned with the common attributes or relations between any three numbers,
and not with their absolute magnitude – for example, they are not rules like ``all numbers above (or below) 50.''\\

Your goal is to discover both rules: the rule for DAX triples and the rule for MED triples.
You will then be told if the triple conforms with the DAX rule or the MED rule.\\

At each turn, you may perform one of two actions:

- Test: propose a new set of three integers, denoted [a,b,c]. You will then be told whether that triple conforms with the DAX or MED rule (``DAX'' or ``MED'').\\
- Announce: based on all previous tests, state what you believe the two hidden rules are. You will not receive any feedback for announcements.\\

Each instruction you receive will specify what to do.\\

Format (must follow exactly):

- If the instruction is Turn - Announce, output exactly two lines:  \\
  Announce: DAX rule - \textless one short sentence \textgreater  \\
  Announce: MED rule - \textless one short sentence \textgreater\\

- If the instruction is Turn - Test, output exactly one line:  \\
  Check: [a,b,c]\\

A DAX triple is: [12, -36, -36].  \\
Let's begin.  \\
Turn - Announce.
\end{PromptBlock}

\paragraph{Example interaction (hidden rule: all numbers are even).}
\emph{The following transcript continues directly from the final ``Turn--Announce'' in the initial prompt above.}

\begin{verbatim}
[model] Announce: DAX rule - The second and third number are equal.
Announce: MED rule - The second and third number are not equal.

[environment] Turn - Test

[model] Check: [2, 4, 6]

[environment] DAX. Turn - Announce

[model] Announce: DAX rule - All three numbers are even.
Announce: MED rule - All three numbers are not even.

[environment] Turn - Test

[model] Check: [2, 4, 5]

[environment] MED. Turn - Announce

[model] Announce: DAX rule - All three numbers are even.
Announce: MED rule - All three numbers are not even.
\end{verbatim}

\subsection{Think-in-Opposites Prompt}
\label{appendix:opposites-prompt}

\begin{PromptBlock}
You will be given three numbers which conform to a simple rule that I have in mind.
This rule is concerned with the common attributes or relations between any three numbers,
and not with their absolute magnitude – for example, it is not a rule like ``all numbers above (or below) 50.''\\

Your goal is to discover this hidden rule by proposing sets of three numbers.
You will then be told if the triple conforms with the hidden rule or not.\\

Before proposing each new triple, look at the example and think about one feature it has.
Then imagine a triple that is opposite with respect to that feature, and test whether it still conforms with the rule.
If it does, that feature is probably irrelevant; if it does not, that feature might be crucial.

In short: to discover the rule, test triples that both confirm and contradict your current idea.\\

At each turn, you may perform one of two actions:

- Test: propose a new set of three integers, denoted [a,b,c].
You will then be told whether that triple conforms with the hidden rule (``YES'' or ``NO'').

- Announce: state what you believe the hidden rule is.
Each instruction you receive will specify what to do.\\

Format (must follow exactly):\\
- If the instruction is Turn - Announce, output exactly one line:  
  Announce: \textless one short sentence naming the rule \textgreater

- If the instruction is Turn - Test, output exactly one line:  
  Check: [a,b,c]\\

A triple that conforms with the hidden rule is: [12, -36, -36].  \\
Let's begin.  \\
Turn - Announce.
\end{PromptBlock}

\paragraph{Example interaction (hidden rule: all numbers are even).}
\emph{The following transcript continues directly from the final ``Turn--Announce'' in the initial prompt above.}

\begin{verbatim}
[model] Announce: The second and third number are equal.

[environment] Turn - Test

[model] Check: [2, 4, 6]

[environment] NO. Turn - Announce

[model] Announce: All three numbers are even.

[environment] Turn - Test

[model] Check: [3, 4, 6]

[environment] NO. Turn - Announce

[model] Announce: All three numbers are even.
\end{verbatim}

\paragraph{Format enforcement.}
If a model response did not follow the required format for the current turn,
we automatically re-prompted the model with the format requirement before proceeding;
such retries were not counted as turns and were not included in the conversation history.

\section{Evaluation Implementation Details}
\label{appendix:judge}

This appendix describes the evaluation pipeline used in our experiments,
including decoding parameters, post-processing rules, and the LLM-based
judge components used to compute task success and confirmation bias
metrics. We use two judges: an announcement correctness judge and a
compatibility judge.

\subsection{Decoding Parameters}

We use the following decoding hyperparameters:
temperature = 0.6, top\_p = 0.95, top\_k = 20, and presence penalty = 0.0.
Standard models (non-thinking) generate up to 256 new tokens per turn,
while thinking-mode models generate up to 16K new tokens per turn.
For thinking-mode models, intermediate \texttt{<think>} traces are stripped
from the dialogue history before the next prompt.

\subsection{Post-processing}

All models must follow the exact output protocol
(\texttt{Check:[a,b,c]} or \texttt{Announce: ...}).
Any formatting violation triggers resampling for the same turn, and the
invalid output is not added to the dialogue history.

\subsection{Announcement Correctness Judge}
\label{appendix:announcement-judge}

\paragraph{Purpose.}
This judge determines whether a model’s announced hypothesis is semantically equivalent
to the ground-truth rule. It is used to compute \emph{Task Success} and
\emph{\# Turns till Success}.

\paragraph{Judge model.}
We use \texttt{Llama-3.3-70B-Instruct} with greedy decoding.

\paragraph{Setup.}
At each \texttt{Announce} turn, the judge is given the ground-truth rule, the model’s
announcement, and lightweight rule-specific guidance (\texttt{\{rule\_guidance\}}).
It outputs a binary decision indicating semantic equivalence.

\paragraph{Rule guidance.}
The \texttt{\{rule\_guidance\}} field provides short natural-language \emph{ACCEPT} and
\emph{REJECT} criteria describing equivalent statements and common near-misses
(broader, narrower, or different rule families).

\subsubsection{Correctness Judge Prompt (Baseline / Think-in-Opposites)}
\label{appendix:judge-baseline}

\begin{PromptBlock}
You are a strict evaluator of SHORT RULE DESCRIPTIONS for number triples.\\

You will receive two short natural-language rules:\\
(1) the model's announced rule, and (2) the ground-truth rule.\\
Decide if they describe the same underlying rule.\\

Guidance specific to the ground-truth rule:\\
\{rule\_guidance\}\\

Rules:\\
- Focus on logical meaning, not phrasing.\\
- If rule families differ (e.g., ordering vs. parity) or one scope is broader/narrower, answer NO.\\
- Ignore superficial differences (synonyms, word order).\\

Respond with ONLY one token: YES or NO.\\

ANNOUNCE TEXT:\\
``\{announced\_rule\}''\\

GROUND-TRUTH RULE:\\
``\{ground\_truth\_rule\}''
\end{PromptBlock}

\subsubsection{Correctness Judge Prompt (Dual-Goal)}
\label{appendix:judge-dualgoal}

\begin{PromptBlock}
You are a strict evaluator of SHORT RULE DESCRIPTIONS for number triples.\\

Context:\\
- This is the DAX–MED experiment. DAX is the positive class; MED is ``not DAX''.\\
- You will receive the participant's full Announce line (may contain both DAX and MED).\\
- Your task: EXTRACT ONLY THE DAX RULE and decide if it has the SAME MEANING as the ground-truth DAX rule.\\

How to extract the DAX rule (in order):\\
1) If the text contains ``DAX:'', ``DAX is'', ``A DAX triple is'', or ``The DAX rule is'', take the phrase following that up to the next ";", ".", ",", or the start of a MED clause.\\
2) Otherwise, if there are two clauses separated by ";" or ", and", take the first clause as the DAX rule and ignore the rest.\\
3) Otherwise, treat the whole line as the DAX rule.\\

Guidance specific to the ground-truth rule:\\
\{rule\_guidance\}\\

Rules:\\
- Ignore phrasing/synonyms if meaning is identical.\\
- If scopes differ (broader/narrower) or the family differs, answer NO.\\

Respond with ONLY one token: YES or NO.\\

ANNOUNCE TEXT:\\
``\{announced\_rule\}''\\

GROUND-TRUTH DAX RULE:\\
``\{ground\_truth\_rule\}''
\end{PromptBlock}

\subsubsection{Announcement Judge Examples}
\label{appendix:announcement-examples}

\paragraph{Baseline / Think-in-Opposites.}
Ground-truth rule: All even

Rule guidance (excerpt):
\begin{quote}\small
ACCEPT: ``all even'', ``each number is even''.\\
REJECT: any rule that allows odds, says ``at least one even'', or uses divisibility instead of parity.
\end{quote}

Model announcement: \texttt{All three numbers are even.} \\
Judge output: \texttt{YES}

Near-miss announcement: \texttt{At least one number is even.} \\
Judge output: \texttt{NO}

\paragraph{Dual-Goal (DAX extraction).}
Ground-truth DAX rule: Ascending

Model announcement:
\texttt{Announce: DAX rule {-} strictly increasing;}\texttt{Announce: MED rule {-} not strictly increasing.}

Judge output: \texttt{YES}

\subsection{Compatibility and Incompatibility Judge}
\label{appendix:compatibility-judge}

\paragraph{Purpose.}
This judge classifies each tested triple as compatible or incompatible with the model’s
current hypothesis, enabling computation of the incompatible-to-compatible ratio (I:C).

\paragraph{Judge model.}
We use \texttt{Qwen-2.5-Coder-32B-Instruct} with decoding temperature 0.2 for stable
rule-to-Python conversion.

\paragraph{Procedure.}
For each \texttt{Test} turn:
\begin{enumerate}
\item The model’s current hypothesis is taken from the immediately preceding \texttt{Announce} turn.
\item The hypothesis is converted into an executable Python function using an LLM.
\item The tested triple $(a,b,c)$ is evaluated using this function.
\item If the function returns \texttt{True}, the test is compatible; otherwise incompatible.
\end{enumerate}

\subsubsection{Compatibility Judge Prompt (Rule-to-Python)}
\label{appendix:judge-hypothesis}

\begin{PromptBlock}
Your task is to write a Python function that determines whether three integers
(a, b, c) satisfy a given natural-language rule.\\

The function should directly implement the logical condition described in the rule.\\
Use basic arithmetic and comparisons. If the rule mentions ``prime'', ``square'', or ``cube'' numbers,
you may define small helper functions (e.g., is\_prime, is\_cube) nested inside the main function.\\
Do not import any modules. Use pure Python only.\\\

Handle negative and zero values safely — especially for divisibility (avoid divide-by-zero errors).\\
Return True if the triple satisfies the rule, otherwise False.\\

---\\
FORMAT \& SAFETY CONSTRAINTS (MANDATORY):\\

- Output ONLY the function definition. No prose, no explanations, no code fences.\\
- Signature must be exactly:\\
  def rule(a: int, b: int, c: int) -> bool:\\
- You may define small helper functions NESTED INSIDE rule().\\
- Do NOT use: import, from, global, nonlocal, with, try/except, raise, class, lambda,
  eval, exec, file or network I/O, or context managers.\\
- Use only pure Python integer arithmetic, comparisons, loops, and these builtins:
  abs, min, max, range, len, all, any, sum.\\
- You may only call functions that are:\\
  * the parameters a, b, c,\\
  * local variables or helper functions defined inside rule, or\\
  * the allowed builtins listed above.\\

  If a name is not in this list, you MUST NOT use it.\\
  In particular, do NOT use: sorted, set, list, dict, pow, divmod, enumerate, zip, map, filter,
  or any other builtin.\\

- Handle negatives and zero safely; avoid division/modulo by zero.\\

- Before you output the function, silently check that you did not use any disallowed syntax
  and did not call any builtin other than the allowed ones above.\\
  Then output ONLY the final function.\\

---\\

Rule:\\
\{HYPOTHESIS\}
\end{PromptBlock}

\paragraph{Repair loop.}
If the generated function fails to execute (e.g., syntax errors or division-by-zero),
the full error message is returned to the model and a corrected function is requested.
This loop continues until a valid executable function is produced.

\subsubsection{Compatibility Judge Examples}
\label{appendix:compatibility-examples}

\noindent\textbf{Natural-language hypothesis.}
\begin{quote}
\texttt{The gaps strictly increase: (b-a) < (c-b) and both are positive.}
\end{quote}

\noindent\textbf{Generated function.}
\begin{center}
\begin{minipage}{0.8\linewidth}
\begin{verbatim}
def rule(a: int, b: int, c: int) -> bool:
    return (b - a) > 0 and (c - b) > 0 and (b - a) < (c - b)
\end{verbatim}
\end{minipage}
\end{center}

\noindent\textbf{Compatibility evaluation.}
\begin{quote}
\((1,2,4)\) $\rightarrow$ \texttt{compatible}, \quad
\((1,3,5)\) $\rightarrow$ \texttt{incompatible}.
\end{quote}

\subsection{Hypothesis Agreement Analysis}
\label{appendix:hypothesis-agreement}

Our I:C metric treats the model's announced rule as its current hypothesis. 
To assess whether a single announcement reliably reflects the model's hypothesis, 
we perform a resampling analysis on 100 announcements from Qwen3-8B in thinking 
mode across interaction settings. For each announcement, we resample the model 
10 times from the same conversation history and use Gemini-2.5-Pro to determine 
whether each resampled hypothesis agrees with the original announcement.

Across the 100 examples, the average agreement rate is 81.6\%. Moreover, for 
88 of the 100 examples, the original announcement agrees with the majority of 
the resampled hypotheses. These results suggest that the announced rule generally 
reflects the model's dominant hypothesis.

\section{Full Model Results}
\label{appendix:full-results}

Tables~\ref{tab:overall-thinking-aligned} and~\ref{tab:overall-nonthinking-aligned} report the full results for all models. Column definitions are provided below.

\vspace{0.3em}
\noindent\textbf{Column definitions.}
\begin{itemize}
    \item \textbf{Task Success}: Fraction of episodes in which the model eventually guessed the hidden rule correctly.
    
    \item \textbf{First Guess}: Fraction of episodes in which the model’s initial hypothesis was correct, prior to any testing.
    
    \item \textbf{\# Turns till Succ.}: Average number of test turns performed before the first correct hypothesis, computed over solved episodes only.
    
    \item \textbf{I:C$_{\text{sol}}$}: Micro-averaged ratio of incompatible to compatible tests \emph{before the first correct hypothesis}, computed over episodes that were eventually solved.
    
    \item \textbf{I:C$_{\text{uns}}$}: Micro-averaged ratio of incompatible to compatible tests over all test turns in episodes that were not solved.
    
    \item \textbf{I:C$_{\text{all}}$}: Micro-averaged ratio of incompatible to compatible tests aggregated over both solved (up to the first correct hypothesis) and unsolved episodes.
    
    \item \textbf{\# Tkns per Turn (Avg$_{\text{sol}}$)}: Average number of tokens generated per model turn, computed over turns up to the first correct hypothesis in solved episodes.
    
    \item \textbf{\# Tkns per Turn (Avg$_{\text{uns}}$)}: Average number of tokens generated per model turn, computed over all turns in unsolved episodes.
    
    \item \textbf{\# Tkns per Turn (Avg$_{\text{all}}$)}: Average number of tokens generated per model turn, aggregated over both solved and unsolved episodes.
\end{itemize}

\begin{table}[!h]
\centering
\small
\caption{Results across all rules and episodes for Thinking Models.}
\label{tab:overall-thinking-aligned}
\setlength{\tabcolsep}{1.5pt}
\begin{tabular}{l l c c c c c c c c c}
\toprule
Model & Variant &
Task &
First &
\# Turns until &
I:C$_{\text{sol}}$ &
I:C$_{\text{uns}}$ &
I:C$_{\text{all}}$ &
\multicolumn{3}{c}{\# Tkns per Turn} \\
\cmidrule(lr){9-11}
& & Suc. $\uparrow$  & Guess & Suc. $\downarrow$  & $\uparrow$ & $\uparrow$ & $\uparrow$ &
Avg$_{\text{sol}}$ &
Avg$_{\text{uns}}$ &
Avg$_{\text{all}}$ \\
\midrule

Qwen3-8B & Baseline
& 0.212 & 0.087 & 10.875 & 0.101 & 0.150 & 0.147
& 958.2 & 1123.2 & 1088.2 \\
& Dual-Goal
& 0.400 & 0.075 & 7.365  & 0.813 & 0.525 & 0.546
& 1763.2 & 2394.1 & 2141.7 \\
& Think-in-Opposites
& 0.363 & 0.000 & 9.008  & 0.826 & 0.314 & 0.353
& 1547.9 & 1811.9 & 1716.1 \\
\midrule

Qwen3-14B & Baseline
& 0.338 & 0.125 & 7.393  & 0.130 & 0.192 & 0.186
& 991.5 & 1274.1 & 1178.6 \\
& Dual-Goal
& 0.575 & 0.138 & 7.661  & 0.609 & 0.369 & 0.404
& 1029.1 & 1411.8 & 1191.7 \\
& Think-in-Opposites
& 0.475 & 0.125 & 7.082  & 0.435 & 0.243 & 0.262
& 1198.5 & 1453.6 & 1332.4 \\
\midrule

Qwen3-32B & Baseline
& 0.413 & 0.150 & 10.833 & 0.235 & 0.131 & 0.141
& 842.9 & 1161.4 & 1029.8 \\
& Dual-Goal
& 0.600 & 0.125 & 10.287 & 0.780 & 0.771 & 0.773
& 1191.8 & 1893.0 & 1472.3 \\
& Think-in-Opposites
& 0.613 & 0.125 & 6.381  & 1.092 & 0.457 & 0.544
& 1397.9 & 1532.4 & 1450.0 \\
\midrule

DeepSeek-R1 & Baseline
& 0.350 & 0.075 & 8.103  & 0.416 & 0.315 & 0.321
& 830.2 & 1026.3 & 957.6 \\
-Llama-70B& Dual-Goal
& 0.500 & 0.075 & 8.385  & 0.725 & 0.588 & 0.608
& 1150.9 & 1271.4 & 1211.2 \\
& Think-in-Opposites
& 0.463 & 0.107 & 10.073  & 0.596 & 0.461 & 0.480
& 998.6 & 1136.7 & 1072.8 \\
\midrule

QwQ-32B & Baseline
& 0.500 & 0.113 & 11.752 & 0.657 & 0.313 & 0.362
& 1269.5 & 2245.5 & 1757.5 \\
& Dual-Goal
& 0.575 & 0.100 & 10.755 & 1.065 & 0.753 & 0.814
& 1679.5 & 3085.8 & 2277.2 \\
& Think-in-Opposites
& 0.588 & 0.088 & 10.282 & 0.913 & 0.269 & 0.371
& 1386.9 & 1575.3 & 1464.5 \\
\midrule

Gemini-2.5-Pro & Baseline
& 0.725 & 0.087 & 11.136 & 0.849 & 0.564 & 0.656
& -- & -- & --\\
& Dual-Goal
& 0.775 & 0.113 & 10.098 & 0.819 & 0.734 & 0.768
& -- & -- & -- \\
& Think-in-Opposites
& 0.775 & 0.062 & 8.520 & 1.069 & 0.585 & 0.748
& -- & -- & -- \\

\bottomrule
\end{tabular}
\end{table}

\begin{table}[!t]
\centering
\small
\caption{Results across all rules and episodes for Non-thinking Models.}
\label{tab:overall-nonthinking-aligned}
\setlength{\tabcolsep}{1.5pt}
\begin{tabular}{l l c c c c c c c c c}
\toprule
Model & Variant &
Task &
First &
\# Turns until &
I:C$_{\text{sol}}$ &
I:C$_{\text{uns}}$ &
I:C$_{\text{all}}$ &
\multicolumn{3}{c}{\# Tkns per Turn} \\
\cmidrule(lr){9-11}
& & Suc. $\uparrow$  & Guess & Suc. $\downarrow$  & $\uparrow$ & $\uparrow$ & $\uparrow$ &
Avg$_{\text{sol}}$ &
Avg$_{\text{uns}}$ &
Avg$_{\text{all}}$ \\
\midrule

Qwen3-8B & Baseline
& 0.062 & 0.013 & 5.000
& 0.174 & 0.354 & 0.353
& 8.5 & 8.3 & 8.4 \\
& Dual-Goal
& 0.075 & 0.075 & 0.000
& -- & 0.574 & 0.574
& 21.0 & 12.7 & 13.4 \\
& Think-in-Opposites
& 0.062 & 0.000 & 15.250
& 0.171 & 0.298 & 0.295
& 7.3 & 8.5 & 8.4 \\
\midrule

Qwen3-14B & Baseline
& 0.225 & 0.175 & 3.000
& 0.600 & 0.724 & 0.723
& 3.8 & 8.7 & 7.6 \\
& Dual-Goal
& 0.125 & 0.113 & 0.500
& 0.000 & 0.411 & 0.411
& 13.7 & 15.2 & 15.0 \\
& Think-in-Opposites
& 0.287 & 0.163 & 4.120
& 2.211 & 0.878 & 0.896
& 4.1 & 8.1 & 7.0 \\
\midrule

Qwen3-32B & Baseline
& 0.325 & 0.225 & 4.933
& 0.544 & 0.494 & 0.495
& 5.0 & 7.3 & 6.6 \\
& Dual-Goal
& 0.100 & 0.100 & 0.000
& -- & 0.456 & 0.456
& 19.9 & 14.9 & 15.4 \\
& Think-in-Opposites
& 0.362 & 0.200 & 3.760
& 0.441 & 0.641 & 0.631
& 5.4 & 6.5 & 6.1 \\
\midrule

LLaMA-3.3- & Baseline
& 0.212 & 0.050 & 9.370
& 0.382 & 0.368 & 0.369
& 14.2 & 14.0 & 14.1 \\
70B-Instruct& Dual-Goal
& 0.200 & 0.025 & 12.869
& 0.709 & 0.521 & 0.529
& 17.8 & 25.1 & 23.7 \\
& Think-in-Opposites
& 0.375 & 0.062 & 10.026
& 0.699 & 0.478 & 0.499
& 12.7 & 15.9 & 14.7 \\
\midrule

GPT-4o & Baseline
& 0.113 & 0.113 & 0.000
& -- & 0.246 & 0.246
& 7.5 & 10.8 & 10.4 \\
& Dual-Goal
& 0.138 & 0.100 & 6.750
& 0.265 & 0.342 & 0.341
& 12.8 & 17.8 & 17.1 \\
& Think-in-Opposites
& 0.163 & 0.100 & 2.500
& 0.097 & 0.167 & 0.166
& 9.1 & 11.0 & 10.7 \\

\bottomrule
\end{tabular}
\end{table}

\subsection{Results by Rule Category}
\label{appendix:rule-category}

We further break down performance by rule category for thinking models. Table~\ref{tab:rule-category} reports Task Success and I:C for ordering, parity, sign, and number-property rules. In most cases, moving from Baseline to Dual-Goal or Think-in-Opposites (TiO) increases both Task Success and I:C, suggesting that the intervention effects are not concentrated in a single rule category.

\begin{table}[!t]
\centering
\small
\caption{\textbf{Wason task performance by rule category for thinking models.}
We report Task Success and I:C for each interaction setting across ordering, parity, sign, and number-property rules.}
\label{tab:rule-category}
\setlength{\tabcolsep}{1.5pt}
\begin{tabular}{l l c c c c c c c c}
\toprule
Model & Variant &
\multicolumn{2}{c}{Ordering} &
\multicolumn{2}{c}{Parity} &
\multicolumn{2}{c}{Sign} &
\multicolumn{2}{c}{Number Properties} \\
\cmidrule(lr){3-4}
\cmidrule(lr){5-6}
\cmidrule(lr){7-8}
\cmidrule(lr){9-10}
& &
Task Suc. & I:C &
Task Suc. & I:C &
Task Suc. & I:C &
Task Suc. & I:C \\
\midrule

DeepSeek-R1- & Baseline
& 0.160 & 0.29
& 0.133 & 0.27
& 0.067 & 0.22
& 0.640 & 0.53 \\
Llama-70B & Dual-Goal
& 0.360 & 0.67
& 0.333 & 0.63
& 0.600 & 0.39
& 0.520 & 0.63 \\
& Think-in-Opposites
& 0.360 & 0.63
& 0.333 & 0.48
& 0.067 & 0.25
& 0.680 & 0.62 \\
\midrule

Gemini-2.5-Pro & Baseline
& 0.600 & 0.66
& 0.867 & 0.75
& 0.400 & 0.51
& 0.920 & 0.83 \\
& Dual-Goal
& 0.600 & 0.79
& 0.533 & 0.88
& 0.933 & 0.52
& 0.880 & 0.84 \\
& Think-in-Opposites
& 0.720 & 0.80
& 0.933 & 1.05
& 0.400 & 0.58
& 0.960 & 0.98 \\
\midrule

Qwen3-8B & Baseline
& 0.080 & 0.13
& 0.000 & 0.15
& 0.000 & 0.08
& 0.480 & 0.23 \\
& Dual-Goal
& 0.320 & 0.49
& 0.267 & 0.74
& 0.533 & 0.83
& 0.320 & 0.44 \\
& Think-in-Opposites
& 0.240 & 0.39
& 0.400 & 0.55
& 0.133 & 0.29
& 0.440 & 0.32 \\
\midrule

Qwen3-14B & Baseline
& 0.280 & 0.24
& 0.067 & 0.23
& 0.067 & 0.03
& 0.520 & 0.24 \\
& Dual-Goal
& 0.400 & 0.30
& 0.267 & 0.27
& 0.733 & 0.27
& 0.640 & 0.71 \\
& Think-in-Opposites
& 0.360 & 0.31
& 0.333 & 0.34
& 0.133 & 0.17
& 0.680 & 0.29 \\
\midrule

Qwen3-32B & Baseline
& 0.360 & 0.14
& 0.067 & 0.16
& 0.267 & 0.09
& 0.560 & 0.21 \\
& Dual-Goal
& 0.480 & 0.62
& 0.400 & 0.60
& 0.800 & 0.85
& 0.560 & 1.04 \\
& Think-in-Opposites
& 0.520 & 0.52
& 0.467 & 0.59
& 0.533 & 0.23
& 0.760 & 0.92 \\
\midrule

QwQ-32B & Baseline
& 0.400 & 0.39
& 0.400 & 0.25
& 0.067 & 0.14
& 0.760 & 0.93 \\
& Dual-Goal
& 0.560 & 0.67
& 0.467 & 0.72
& 0.733 & 0.57
& 0.520 & 1.53 \\
& Think-in-Opposites
& 0.560 & 0.40
& 0.533 & 0.31
& 0.267 & 0.19
& 0.760 & 0.67 \\

\bottomrule
\end{tabular}
\end{table}

\section{Distillation Training Details}
\label{appendix:distillation-details}

This section provides implementation details for the context distillation experiments
in Section~\ref{sec:distillation}.

\subsection{Training Data}

We train on \textbf{all \texttt{Test} turns} produced by the teacher, using \emph{untrimmed}
trajectories (including unsuccessful episodes and all turns after the first correct announcement).
Each training example consists of the dialog history up to a \texttt{Test} turn (formatted with the
\emph{baseline} prompt) and the teacher's next-turn response (generated under the \emph{intervention} prompt).
We use the same train/validation/test rule splits as in Appendix~\ref{appendix:data}.

\subsection{Learning-rate Sweeps and Selected Checkpoints}
\label{appendix:distill-sweeps}
We fine-tune students with LoRA-based supervised fine-tuning.

\paragraph{Other fixed hyperparameters.}
For Qwen3-8B students, we use an effective batch size 64,
epochs 5, LoRA rank $r{=}16$ ($\alpha{=}32$), warmup ratio 0.05, and weight decay 0.0.
For the Qwen3-32B student, we use an effective batch size of 64,
epochs 5, LoRA rank $r{=}32$ ($\alpha{=}64$), warmup ratio 0.05, and weight decay 0.0.

\paragraph{Checkpoint selection.}
Within each teacher--student setting, we select the checkpoint with lowest validation loss and use it for the evaluations reported in the main paper. Table~\ref{tab:distill-sweeps} summarizes the learning-rate sweep ranges explored for each teacher--student pair and the selected checkpoint used in all reported evaluations.

\begin{table}[htbp]
\centering
\small
\setlength{\tabcolsep}{3pt}
\renewcommand{\arraystretch}{1.15}
\caption{Learning-rate sweeps and selected checkpoints for distillation.}
\label{tab:distill-sweeps}
\begin{tabular}{@{}p{0.38\linewidth} p{0.28\linewidth} c c@{}}
\toprule
\textbf{Student $\leftarrow$ Teacher}
& \textbf{LR sweep range}
& \textbf{Best LR}
& \textbf{Best step (epoch)} \\
\midrule

Qwen3-8B $\leftarrow$ Qwen3-8B (Think-in-Opposites)
& $[1{\times}10^{-5},\,5{\times}10^{-4}]$ (7 values)
& $7.5{\times}10^{-5}$
& 3400 (3.02) \\

Qwen3-8B $\leftarrow$ Qwen3-32B (Baseline)
& $[1{\times}10^{-4},\,2{\times}10^{-3}]$ (6 values)
& $2{\times}10^{-4}$
& 3200 (2.84) \\

Qwen3-8B $\leftarrow$ Qwen3-32B (Think-in-Opposites)
& $[1{\times}10^{-5},\,5{\times}10^{-4}]$ (6 values)
& $2{\times}10^{-4}$
& 4400 (3.91) \\

Qwen3-32B $\leftarrow$ Qwen3-32B (Think-in-Opposites)
& $[2{\times}10^{-6},\,5{\times}10^{-5}]$ (6 values)
& $5{\times}10^{-5}$
& 2200 (1.96) \\

\bottomrule
\end{tabular}
\end{table}

\paragraph{Compute.}
All models are fine-tuned on H200 GPUs (140GB each). Qwen3-8B students 
use 1 GPU and Qwen3-32B students use 2 GPUs, with training runs 
taking 2 to 4 days depending on validation loss convergence.

\section{Blicket Task}
\label{appendix:blicket}

\subsection{Original Experiment vs. Our Adaptation}
\label{appendix:blicket-adaptation}

The original Blicket detector experiment~\citep{gopnik2000detecting} is a developmental psychology paradigm designed to study human's causal reasoning. In the setup, humans are presented with \textbf{four objects} and a machine that activates according to a hidden causal rule. \textbf{Two of the objects are designated as blickets}, meaning they causally activate the machine. The hidden rule is either a conjunctive(AND) or disjunctive(OR) combination of the two blickets. Children observe experimenter-provided evidence and, in some conditions, are allowed to intervene by placing objects on the machine themselves. The interaction is flexible and does not follow a fixed turn structure. Participants have to identify the objects that are blickets, though are not required to explicitly state the hidden rule.

Our adaptation diverges from this original setting in two principal ways.

\paragraph{Structured interaction.}
We impose a fixed 45-turn alternating \texttt{Guess}–\texttt{Test} protocol, mirroring our structured adaptation of Wason’s 2–4–6 rule-discovery task. On each \texttt{Test} turn, the model proposes a subset of objects and receives deterministic ON/OFF feedback. On each \texttt{Guess} turn, the model must explicitly announce both (i) the set of objects that are blickets and (ii) the underlying hidden rule. This structured format differs from the flexible exploratory interaction in the original experiment and enables systematic measurement of confirmation bias via the incompatible-to-compatible (I:C) test ratio.

\paragraph{Expanded data design.}
Whereas the original paradigm uses a small object set (four objects in the canonical setup) with a limited causal structure, we expand the hypothesis space along multiple dimensions. Similar to how we broadened Wason’s single hidden ascending rule into a diverse rule family, we increase complexity in the Blicket setting by:
\begin{itemize}
    \item varying the number of objects ($N \in \{4, 8\}$),
    \item varying the number of blickets ($K \in \{2, 3\}$),
    \item introducing an additional XOR rule alongside AND and OR.
\end{itemize}

\subsection{Dataset Statistics}
\label{appendix:blicket-dataset}
Across all configurations, we generate 192 evaluation episodes. 
An \emph{episode} is defined as a single environment instance with:
(i) a fixed set of objects,
(ii) a designated subset of blickets,
(iii) a hidden logical rule (AND, OR, or XOR),
and (iv) a 45-turn alternating Guess–Test interaction.

For each combination of object count ($N$), blicket count ($K$), and rule type, we generate 16 distinct episodes by varying:
\begin{itemize}
    \item the assignment of which objects are designated as blickets, and
    \item the initial object placement shown at the start of the interaction.
\end{itemize}

This results in a balanced data design summarized in Table~\ref{tab:blicket-stats}, yielding
$2 \times 2 \times 3 \times 16 = 192$ evaluation episodes.

\begin{table}[h]
\centering
\small
\caption{Blicket evaluation dataset.}
\label{tab:blicket-stats}
\begin{tabular}{l c c c}
\toprule
Objects ($N$) & Blickets ($K$) & Rule & \# Episodes \\
\midrule
4 & 2 & AND & 16 \\
4 & 2 & OR  & 16 \\
4 & 2 & XOR & 16 \\
\midrule
4 & 3 & AND & 16 \\
4 & 3 & OR  & 16 \\
4 & 3 & XOR & 16 \\
\midrule
8 & 2 & AND & 16 \\
8 & 2 & OR  & 16 \\
8 & 2 & XOR & 16 \\
\midrule
8 & 3 & AND & 16 \\
8 & 3 & OR  & 16 \\
8 & 3 & XOR & 16 \\
\midrule
\multicolumn{3}{r}{Total} & 192 \\
\bottomrule
\end{tabular}
\end{table}

\subsection{Task Instruction Prompts and Examples}
\label{appendix:blicket-prompt}
Below we provide the task instruction prompts used to initialize each episode of the Blicket task.
As in the Wason setup, these instructions are provided once at the beginning of an interaction, after which the model alternates between \texttt{Announce} and \texttt{Test} turns. Across episodes, the initial object configuration varies (i.e., which objects are placed on the device), and the device may initially be either ON or OFF.

We evaluate two variants: a baseline prompt and a Think-in-Opposites (TiO) prompt. 
The TiO version augments the baseline instructions with an explicit strategy encouraging the model to test opposing features of its current hypothesis, adapted to the object-combination setting.

\subsubsection{Baseline prompt}

\begin{PromptBlock}
You are playing a game with several objects and a device.\\

Placing a set of objects on the device causes it to turn on.
This depends on which objects are placed on the device, according to a hidden rule.\\

Your goal is to discover:\\
- which objects are relevant to turn the device on\\
- the hidden rule that determines when the device turns on.\\

At each turn, you will be instructed to do one of two things:

- Test: propose a set of objects to place on the device. You will then receive feedback on the device's state: ON or OFF.\\
- Announce: based on all previous tests, state which objects you believe are relevant to turn the device on and describe the hidden rule. You will not receive any feedback for announcements.\\

Each instruction you receive will specify what to do.\\

Format (must follow exactly):

- If the instruction is Turn - Announce, output exactly one line:\\
  Announce: relevant=[object A, object B, object C]; rule=\textless one short description of the rule \textgreater \\
  (there can be any number of relevant objects)

- If the instruction is Turn - Test, output exactly one line:\\
  Test: [object A, object B, object C]\\
  (you can test any number and combination of objects)\\

Important rules:\\
- Always refer to objects using the exact name ``object N''.\\
- Do not include any extra text.\\

--------------------------------------------------\\
Initial configuration (this is not the result of a test):\\

You are in a room. You see a device at the center of this room.

There are also 4 objects scattered around the room. You observe them:\\
object 0 is on the floor, object 1 is on the device,
object 2 is on the floor, object 3 is on the floor.

The device is off. You wonder if there is a relationship between the objects and the device.

--------------------------------------------------\\
Turn - Announce
\end{PromptBlock}

\paragraph{Example interaction (blickets: object 0 and object 1, hidden rule: AND}
\emph{The following transcript continues directly from the final ``Turn--Announce'' in the initial prompt above.}

\begin{verbatim}
[model] Announce: relevant=[object 0, object 2, object 3];
        rule=at least two of these objects must be on the device

[environment] Turn - Test

[model] Test: [object 0, object 1, object 2]

[environment] ON. Turn - Announce

[model] Announce: relevant=[object 0, object 2, object 3];
        rule=at least two of these objects must be on the device

[environment] Turn - Test

[model] Test: [object 0, object 1, object 3]

[environment] ON. Turn - Announce

[model] Announce: relevant=[object 0, object 2, object 3]; 
        rule=at least two of these objects must be on the device
\end{verbatim}

\subsubsection{Think-in-Opposites prompt}
\begin{PromptBlock}
You are playing a game with several objects and a device.\\

Placing a set of objects on the device causes it to turn on.
This depends on which objects are placed on the device, according to a hidden rule.\\

Your goal is to discover:\\
- which objects are relevant to turn the device on\\
- the hidden rule that determines when the device turns on.\\

At each turn, you will be instructed to do one of two things:

- Test: propose a set of objects to place on the device. You will then receive feedback on the device's state: ON or OFF.\\
- Announce: based on all previous tests, state which objects you believe are relevant to turn the device on and describe the hidden rule. You will not receive any feedback for announcements.\\

Each instruction you receive will specify what to do.\\

Strategy: At every test step, follow this reasoning procedure:\\
1. Form a concise current hypothesis of the relevant objects and hidden rule.\\
2. Identify one feature implicit in that hypothesis.\\
3. Test the OPPOSITE of that feature.\\
4. If the hypothesis still holds, the feature is probably irrelevant; if it does not, that feature might be crucial.\\
5. In short, prefer tests that both confirm and contradict your current idea.\\

Format (must follow exactly):

- If the instruction is Turn - Announce, output exactly one line:\\
  Announce: relevant=[object A, object B, object C]; rule=\textless one short description of the rule \textgreater \\
  (there can be any number of relevant objects)

- If the instruction is Turn - Test, output exactly one line:\\
  Test: [object A, object B, object C]\\
  (you can test any number and combination of objects)\\

Important rules:\\
- Always refer to objects using the exact name ``object N''.\\
- Do not include any extra text.\\

--------------------------------------------------\\
Initial configuration (this is not the result of a test):\\

You are in a room. You see a device at the center of this room.

There are also 4 objects scattered around the room. You observe them:\\
object 0 is on the floor, object 1 is on the device,
object 2 is on the floor, object 3 is on the floor.

The device is off. You wonder if there is a relationship between the objects and the device.

--------------------------------------------------\\
Turn - Announce
\end{PromptBlock}

\paragraph{Example interaction (blickets: object 0 and object 1, hidden rule: AND}
\emph{The following transcript continues directly from the final ``Turn--Announce'' in the initial prompt above.}

\begin{verbatim}
[model] Announce: relevant=[object 0, object 2, object 3]; 
        rule=at least two of these objects must be on the device

[environment] Turn - Test

[model] Test: [object 0, object 1]

[environment] ON. Turn - Announce

[model] Announce: relevant=[object 0, object 2, object 3];
        rule=at least one of these objects must be on the device

[environment] Turn - Test

[model] Test: [object 1, object 2, object 3]

[environment] OFF. Turn - Announce

[model] Announce: relevant=[object 0, object 1];
        rule=both of these objects must be on the device
\end{verbatim}

\subsection{Judge Prompts}
\label{appendix:blicket-judge-prompts}

\subsubsection{Correctness Judge Prompt}
\begin{PromptBlock}
Decide whether the agent's ANNOUNCEMENT is COMPLETELY CORRECT.\\

Ground truth:\\
- True relevant objects: \{true\_blickets\}\\
- True rule type: \{true\_rule\}\\

Rule semantics:\\
- If true\_rule\_type is 'disjunctive', the device turns on iff at least ONE of the relevant objects is on the device.\\
- If true\_rule\_type is 'conjunctive', the device turns on iff ALL of the relevant objects are on the device.\\
- If true\_rule\_type is 'xor', the device turns on iff EXACTLY ONE of the relevant objects is on the device (not zero, not two+).\\

Agent announcement (verbatim):\\
\{announce\_text\}\\

Mark True ONLY IF BOTH conditions hold:\\

1) The set of guessed relevant objects matches the true relevant objects exactly
   (same members; order doesn't matter; no extras; no missing).\\
   - The agent may say "relevant objects are ..." OR directly name the correct objects; both are acceptable.\\

2) The stated rule matches the true rule semantics exactly.\\
   - If the agent states OR / ANY / AT LEAST ONE when the true rule is conjunctive, that is False.\\
   - If the agent states AND / ALL / NEEDS BOTH when the true rule is disjunctive, that is False.\\
   - If the agent states XOR / EXACTLY ONE / A OR B BUT NOT BOTH / ONE AND ONLY ONE when the true rule is xor, that is True.\\
   - If the agent states relevant object A on and relevant object B off, but does not specify the other direction (A off, B on) when the true rule is xor, that is False.\\
   - If the agent states odd number of relevant objects on when there are 3 or more relevant objects and the true rule is xor, that is False.\\

If there is any ambiguity, missing detail, partial correctness, or mismatch, output False.\\

Output exactly one word: True or False.
\end{PromptBlock}

\paragraph{Example}

Ground-truth rule: AND; 
Blickets = object 0 and object 1

Model announcement: \texttt{relevant=[object 0, object 1]; rule=both the relevant objects must be on the device} \\
Judge output: \texttt{YES}

Near-miss announcement: \texttt{relevant=[object 0, object 1]; rule=any of the relevant objects must be on the device} \\
Judge output: \texttt{NO}

\subsubsection{Compatibility Judge Prompt}
\begin{PromptBlock}
Your task is to write a Python function that determines whether a given device would be ON,
according to a natural-language hypothesis about relevant objects and a rule.\\

The input is a list named state of length num\_objects.\\
- state[i] is True iff ``object i'' is on the device.\\
- state[i] is False iff ``object i'' is NOT on the device.\\

The function should implement the ON-condition described by the hypothesis.\\
Return True if the hypothesis predicts the device is ON for the given state, otherwise False.\\

You may assume objects are named ``object 0'' ... ``object (num\_objects-1)''.\\
Empty sets are allowed.\\

---\\
FORMAT \& SAFETY CONSTRAINTS (MANDATORY):\\

- Output ONLY the function definition. No prose, no explanations, no code fences.\\
- Signature must be exactly:\\
  def rule(state: list) -\textgreater bool:\\
- You may define small helper functions NESTED INSIDE rule().\\
- Do NOT use: import, from, global, nonlocal, with, try/except, raise, class, lambda,
  eval, exec, file or network I/O, or context managers.\\
- Use only pure Python booleans, indexing into state, comparisons, if/else, loops.\\
- Allowed builtins: abs, min, max, range, len, all, any, sum, int, bool, isinstance.\\
- You may only access:\\
  * the parameter state,\\
  * local variables / helper functions defined inside rule,\\
  * the allowed builtins listed above.\\
- Before you output the function, silently check you did not use disallowed syntax or names.
  Then output ONLY the final function.\\

---\\
Hypothesis:\\
\{HYPOTHESIS\}\\

num\_objects=\{NUM\_OBJECTS\}
\end{PromptBlock}

\noindent\textbf{Example Natural-language hypothesis.}
\begin{quote}
\texttt{relevant=[object 0, object 2, object 3]; rule=at least two of these objects must be on the device}
\end{quote}

\noindent\textbf{Generated function.}
\begin{center}
\begin{minipage}{0.8\linewidth}
\begin{verbatim}
def rule(state: list) -> bool:
    count = 0
    if state[0]:
        count += 1
    if state[2]:
        count += 1
    if state[3]:
        count += 1
    return count >= 2
\end{verbatim}
\end{minipage}
\end{center}

\noindent\textbf{Compatibility evaluation.}
\begin{quote}
\((object 0, object 1)\) $\rightarrow$ \texttt{incompatible}, \quad
\((object 0, object 1, object2)\) $\rightarrow$ \texttt{compatible}.
\end{quote}

\section{Distillation Results}
\subsection{Wason Distillation Results}
\label{appendix:wason-distillation-results}
\begin{table}[H]
\centering
\small
\setlength{\tabcolsep}{4pt}
\renewcommand{\arraystretch}{1.1}
\caption{
Model performance after distillation. We mark statistically significant improvements over the base model with superscript $^*$.
}
\label{tab:distillation}
\begin{tabular}{l l c c c}
\toprule
\textbf{Student (+ Teacher)} & \textbf{Prompt} & \textbf{Task Suc.} & \textbf{\# Turns until Suc.} & \textbf{I:C} \\
\midrule

Qwen3-8B & -- & 0.21 & 10.88 & 0.15 \\
Qwen3-8B & TiO & 0.36 & 7.37 & 0.35 \\

Qwen3-32B & -- & 0.41 & 10.83 & 0.14 \\
Qwen3-32B & TiO & 0.61 & 6.38 & \textbf{0.54} \\

\midrule
\multicolumn{5}{l}{\textbf{Intervention Distillation}} \\

Qwen3-8B + 8B TiO & -- & 0.35 & 11.92 & 0.31$^*$ \\
Qwen3-32B + 32B TiO & -- & \textbf{0.64}$^*$ & 9.55 & 0.33$^*$ \\

\midrule
\multicolumn{5}{l}{\textbf{Cross-scale Distillation}} \\

Qwen3-8B + 32B & -- & 0.44$^*$ & 14.44 & 0.22 \\

\midrule
\multicolumn{5}{l}{\textbf{Cross-scale \& Intervention Distillation}} \\

Qwen3-8B + 32B TiO & -- & 0.59$^*$ & 11.41 & 0.38$^*$ \\

\bottomrule
\end{tabular}
\end{table}

\subsection{Blicket Transfer Results}
\label{appendix:blicket-transfer-results}
\begin{table}[H]
\centering
\small
\setlength{\tabcolsep}{4pt}
\renewcommand{\arraystretch}{1.1}
\caption{
Performance of distilled models trained on the Wason task, evaluated on the Blicket test. Statistically significant improvements over the base model are marked with $^*$.
}
\label{tab:blicket-performance}
\begin{tabular}{l l c c c}
\toprule
\textbf{Student (+ Teacher)} & \textbf{Prompt} & \textbf{Task Suc.} & \textbf{\# Turns until Suc.} & \textbf{I:C} \\
\midrule

Qwen3-8B & Base & 0.32 & 10.26 & 0.44 \\
Qwen3-8B & TiO & 0.40 & 10.73 & 0.57 \\

Qwen3-32B & Base & 0.57 & 12.18 & 0.54 \\
Qwen3-32B & TiO & \textbf{0.86} & 12.98 & \textbf{1.31} \\

\midrule
\multicolumn{5}{l}{\textbf{Models Distilled on Wason Task}} \\

Qwen3-8B + 8B TiO & Base & 0.34 & 13.96 & 0.48 \\
Qwen3-8B + 32B & Base & 0.36 & 10.25 & 0.53 \\
Qwen3-8B + 32B TiO & Base & 0.38 & 11.84 & 0.51 \\
Qwen3-32B + 32B TiO & Base & 0.77$^*$ & 11.02 & 0.88$^*$ \\

\midrule
\multicolumn{5}{l}{\textbf{Models Distilled on Wason Task + TiO Prompt}} \\

Qwen3-8B + 8B TiO & TiO & 0.35 & -- & 0.64 \\
Qwen3-32B + 32B TiO & TiO & 0.85 & -- & 1.35 \\

\bottomrule
\end{tabular}
\end{table}

\subsection{Catastrophic Forgetting Check}
\label{appendix:gsm8k}

Fine-tuning on a narrow task can degrade general capabilities through catastrophic
forgetting~\citep{MCCLOSKEY1989109}. To check whether distillation on the
Wason task harms pre-existing mathematical reasoning ability, we evaluate all
Wason-distilled models on the GSM8K test set~\citep{cobbe2021training} (1{,}319 examples)
alongside their base models. Since our distillation targets exploration behavior in multi-turn setups, we do not expect it to influence single-turn mathematical reasoning. The results below in Table~\ref{tab:gsm8k} confirm this with differences of at most 2\% relative to base models. We see small degradation for the 8B model, but no degradation for the 32B model. Distilling the 32B model into the 8B model for the Wason task even resulted in performance gains.

\begin{table}[H]
\centering
\caption{GSM8K accuracy for base models and their Wason-distilled counterparts.
Across all distilled models, accuracy changes by at most 2\% relative to base models.}
\label{tab:gsm8k}
\begin{tabular}{lc}
\toprule
\textbf{Student (+ Teacher)} & \textbf{GSM8K Acc.} \\
\midrule
Qwen3-8B                  & 0.919 \\
Qwen3-32B                 & 0.930 \\
\midrule
\textbf{Models Distilled with Wason Task}\\
Qwen3-8B + 8B TiO         & 0.899 \\
Qwen3-8B + 32B Base       & 0.937 \\
Qwen3-8B + 32B TiO        & 0.933 \\
Qwen3-32B + 32B TiO       & 0.932 \\
\bottomrule
\end{tabular}
\end{table}

\section{Mathematical Reasoning Experiment Details}
\label{appendix:math-reasoning}

This section provides additional experimental details for the mathematical reasoning evaluation in Section~\ref{sec:math-generalization}.

\subsection{Evaluation Setup}
\label{appendix:math-setup}

We evaluate Qwen3-8B and Qwen3-32B and their corresponding same-scale intervention-distilled models (8B $\leftarrow$ 8B TiO and 32B $\leftarrow$ 32B TiO) on AIME 2025 and Beyond AIME. The distilled models are trained exclusively on the Wason task as described in Section~\ref{sec:distillation}.

For each problem, the model first produces a solution and final answer. If the final answer is incorrect, we retain the full conversation history, inform the model that its previous answer was incorrect, and ask it to review its previous reasoning and try again. We allow up to five retries after the initial attempt, resulting in a maximum of six attempts per problem.

Once a problem is answered correctly, it is considered solved for that attempt and all subsequent attempts. Accordingly, the accuracy reported in Figure~\ref{tab:math-retries} is cumulative: at each attempt, we report the fraction of problems that have been solved at least once by that point.

We use the same retry procedure for the base and distilled models. No Think-in-Opposites instruction or other intervention is provided at inference time.

\subsection{Prompts}
\label{appendix:math-prompts}

For the initial attempt, we use the following prompt:

\begin{PromptBlock}
\small
Solve the following problem. Give concise final working with only the key steps, then write your final answer after \texttt{\#\#\#\#}.

Problem:

\{problem\}
\end{PromptBlock}

After an incorrect answer, the model receives the full conversation history followed by:

\begin{PromptBlock}
\small
Your answer was incorrect. Please try again. Review your previous working, fix the mistake, and keep the corrected working concise. Give concise final working with only the key steps, then write your final answer after \texttt{\#\#\#\#}.
\end{PromptBlock}

The same retry prompt is provided after each incorrect attempt, up to a maximum of five retries.

\section{Statistical Significance Testing}
\subsection{Intervention Effects}
\label{appendix:intervention-stats}

To assess whether intervention prompts improve performance relative to
the baseline prompt, we evaluate statistical significance using
paired permutation tests.

For each metric (Task Success and I:C), we first compute the metric
under the baseline prompt and under the intervention prompt
(Dual-Goal or Think-in-Opposites). Metrics are aggregated across
all episodes. We then compute the observed difference
between intervention and baseline:

\[
\Delta_{\text{obs}} = M_{\text{intervention}} - M_{\text{baseline}} .
\]

To construct a null distribution, we randomly permute the baseline and
intervention labels at the episode level while keeping the pairing
between corresponding episodes. After each permutation, we recompute
the difference between the intervention and baseline conditions.
Repeating this process many times produces a distribution of differences
expected under the null hypothesis that the intervention has no effect.

We perform $50{,}000$ permutations. Let $k$ denote the number of
permuted differences that are greater than or equal to the observed
difference $\Delta_{\text{obs}}$. The $p$-value is computed as

\[
p = \frac{k + 1}{N + 1},
\]

where $N$ is the number of permutations. The $+1$ correction prevents
zero $p$-values and yields an unbiased estimate of significance.

Tables~\ref{tab:intervention-deltas} and \ref{tab:tio-deltas} report
the observed mean change ($\Delta$) relative to the baseline prompt
and the corresponding $p$-values. We further apply Benjamini--Hochberg (BH) correction to control the
false discovery rate across the 8 Wason intervention comparisons
(2 interventions $\times$ 2 metrics $\times$ 2 model types), and
report the resulting $q$-values (BH-adjusted $p$-values).

\begin{table}[!h]
\centering
\small
\setlength{\tabcolsep}{3pt}
\caption{\textbf{Effect of interventions on the Wason Task.}
$\Delta$ denotes the change relative to the baseline prompt.
Statistical significance is assessed using paired permutation tests
over episode pairs ($50{,}000$ permutations). $q$-values are computed
using Benjamini--Hochberg correction over the 8 Wason intervention comparisons.}
\label{tab:intervention-deltas}
\begin{tabular}{l l r r r | l l r r r}
\toprule
\multicolumn{5}{c}{\textbf{Thinking models}} &
\multicolumn{5}{c}{\textbf{Non-thinking models}} \\
\cmidrule(lr){1-5} \cmidrule(lr){6-10}

Metric & Intervention & Mean $\Delta$ & $p$ & $q$ &
Metric & Intervention & Mean $\Delta$ & $p$ & $q$ \\
\midrule

Task  & Dual-Goal          & +0.148 & $<2\times10^{-5}$ & $<4\times10^{-5}$ &
Task  & Dual-Goal          & $-0.060$ & 0.0065 & 0.0087 \\
     Success       & TiO & +0.123 & $<2\times10^{-5}$ & $<4\times10^{-5}$ &
       Success     & TiO & +0.063 & 0.0033 & 0.0053 \\

\addlinespace

I:C         & Dual-Goal          & +0.375 & $<2\times10^{-5}$ & $<4\times10^{-5}$ &
I:C         & Dual-Goal          & +0.048 & 0.2386 & 0.2727 \\
            & TiO & +0.168 & $<2\times10^{-5}$ & $<4\times10^{-5}$ &
            & TiO & +0.020 & 0.5995 & 0.5995 \\

\bottomrule
\end{tabular}
\end{table}

\begin{table}[!h]
\centering
\small
\setlength{\tabcolsep}{2pt}
\renewcommand{\arraystretch}{0.9}

\caption{\textbf{Effect of Think-in-Opposites (TiO) on the Blicket test.}
$\Delta$ denotes the change relative to the baseline prompt.
Statistical significance is assessed using paired permutation tests over episode pairs ($50{,}000$ permutations).
$q$-values are computed using Benjamini--Hochberg correction over the 4 Blicket comparisons.}
\label{tab:tio-deltas}
\begin{tabular}{l r r r | r r r}
\toprule
& \multicolumn{3}{c}{\textbf{Non-thinking}}
& \multicolumn{3}{c}{\textbf{Thinking}} \\
\cmidrule(lr){2-4} \cmidrule(lr){5-7}
\textbf{Metric}
& $\Delta$ & $p$ & $q$
& $\Delta$ & $p$ & $q$ \\
\midrule
Task Success
& +0.003 & 0.936 & 0.936
& +0.092 & $<2\times10^{-5}$ & $<8\times10^{-5}$ \\

I:C
& +0.064 & 0.033 & 0.044
& +0.231 & $<2\times10^{-5}$ & $<8\times10^{-5}$ \\
\bottomrule
\end{tabular}
\end{table}

\subsection{Distillation Effects}
\label{appendix:distillation-stats}

We additionally assess whether the performance gains from intervention
distillation are statistically significant relative to the corresponding
baseline models.

We focus on same-scale intervention distillation settings, where a student
model is trained on data generated using Think-in-Opposites prompts from a
teacher of the same scale (e.g., 8B $\rightarrow$ 8B, 32B $\rightarrow$ 32B).
These settings isolate the effect of distilling intervention-induced reasoning
behavior without confounding scale differences.

Statistical significance is evaluated using the same paired permutation test
procedure described in Section~\ref{appendix:intervention-stats}. For each
episode, we pair the baseline model output with the corresponding distilled
model output, compute the observed difference in Task Success and I:C, and
construct a null distribution by randomly swapping labels within each episode.
We perform $50{,}000$ permutations and compute two-sided $p$-values with the
$(k+1)/(N+1)$ correction.

\begin{table}[!h]
\centering
\small
\setlength{\tabcolsep}{3pt}

\caption{\textbf{Effect of distillation on the Wason Task.}
$\Delta$ denotes the change relative to the corresponding baseline model.
Statistical significance is assessed using paired permutation tests over episode pairs ($50{,}000$ permutations).}
\label{tab:distillation-wason}
\begin{tabular}{l l r r}
\toprule
\textbf{Metric} & \textbf{Model} & \textbf{Mean $\Delta$} & \textbf{$p$} \\
\midrule

Task Success 
& Qwen3-8B + 8B TiO & +0.138 & 0.053 \\
& Qwen3-32B + 32B TiO & +0.225 & 0.00032 \\
& Qwen3-8B + 32B & +0.225 & 0.00078 \\
& Qwen3-8B + 32B TiO & +0.438 & $<2\times10^{-5}$ \\

\addlinespace

I:C 
& Qwen3-8B + 8B TiO & +0.163 & 0.0018 \\
& Qwen3-32B + 32B TiO & +0.185 & $<2\times10^{-5}$ \\
& Qwen3-8B + 32B & +0.077 & 0.184296 \\
& Qwen3-8B + 32B TiO & +0.310 & $<2\times10^{-5}$ \\

\bottomrule
\end{tabular}
\end{table}

\begin{table}[!h]
\centering
\small
\setlength{\tabcolsep}{3pt}

\caption{\textbf{Effect of Wason Distillation on the Blicket Test.}
Models are distilled on the Wason task and evaluated on the Blicket test.
$\Delta$ denotes the change relative to the corresponding baseline model on the Blicket test.
Statistical significance is assessed using paired permutation tests over episode pairs ($50{,}000$ permutations).}
\label{tab:distillation-blicket}
\begin{tabular}{l l r r}
\toprule
\textbf{Metric} & \textbf{Model} & \textbf{Mean $\Delta$} & \textbf{$p$} \\
\midrule

Task Success 
& Qwen3-8B + 8B TiO & +0.021 & 0.674 \\
& Qwen3-32B + 32B TiO & +0.203 & $<2\times10^{-5}$ \\
& Qwen3-8B + 32B Baseline & +0.042 & 0.347 \\
& Qwen3-8B + 32B TiO & +0.063 & 0.150 \\

\addlinespace

I:C 
& Qwen3-8B + 8B TiO & +0.039 & 0.396 \\
& Qwen3-32B + 32B TiO & +0.345 & $<2\times10^{-5}$ \\
& Qwen3-8B + 32B Baseline & +0.092 & 0.051 \\
& Qwen3-8B + 32B TiO & +0.070 & 0.090 \\

\bottomrule
\end{tabular}
\end{table}

\section{Judge Validation}
\label{appendix:judge-validation}

Our evaluation relies on judge LLMs to determine (i) whether a model's
announced rule is correct and (ii) whether a proposed test is compatible with
a previously announced rule. To verify the reliability of these judges, we
conduct a manual audit on a sample of evaluation instances from
both tasks.

For the Wason and Blicket tasks, across all models evaluated on the test set, we randomly sample 200 announcement judgments each (100 labeled correct and
100 labeled incorrect by the judge) and 200 compatibility judgments each (100 labeled compatible and 100
labeled incompatible by the judge). One of the authors manually annotated each sample and this is compared with the decision of the LLM-as-a-judge.

Table~\ref{tab:judge-validation-summary} summarizes the audit results.
Across all four audited settings, the judge decisions achieve $\geq$ 98.5\% agreement with human annotations. Tables~\ref{tab:wason-guess-judge-examples}--
\ref{tab:blicket-compat-judge-examples} show a few representative examples.

\begin{table}[h]
\centering
\small
\setlength{\tabcolsep}{6pt}
\renewcommand{\arraystretch}{1.1}
\caption{Summary of the manual audit for judge validation.}
\begin{tabular}{lccc}
\toprule
\textbf{Setting} & \textbf{Sample size} & \textbf{Pos./Neg.} & \textbf{Agreement} \\
\midrule
Wason announcement     & 200 & 100 / 100 & 198 / 200 (99.0\%) \\
Wason compatibility    & 200 & 100 / 100 & 197 / 200 (98.5\%) \\
Blicket announcement   & 200 & 100 / 100 & 199 / 200 (99.5\%) \\
Blicket compatibility  & 200 & 100 / 100 & 197 / 200 (98.5\%) \\
\bottomrule
\end{tabular}
\label{tab:judge-validation-summary}
\end{table}

\begin{table*}[h]
\centering
\footnotesize
\setlength{\tabcolsep}{4pt}
\renewcommand{\arraystretch}{1.05}
\caption{Representative examples from the manual audit of Wason announcement judgments.}
\begin{tabular}{p{2.5cm} p{6.0cm} p{1.8cm} p{1.8cm}}
\toprule
\textbf{Correct announcement} & \textbf{Model announcement} & \textbf{Judge} & \textbf{Human} \\
\midrule
All end with 1 & All numbers end with the digit 1. & CORRECT & CORRECT \\
Mixed signs & The triple contains at least one negative number and at least one non-negative number. & CORRECT & CORRECT \\
Descending & The numbers are in decreasing order. & CORRECT & CORRECT \\
All cube numbers & The numbers are in a geometric progression with a common ratio of -3. & INCORRECT & INCORRECT \\
All positive & The three numbers form an arithmetic sequence. & INCORRECT & INCORRECT \\
Contains a prime & The sum of the first two numbers is twice the third. & INCORRECT & INCORRECT \\
\bottomrule
\end{tabular}
\label{tab:wason-guess-judge-examples}
\end{table*}

\begin{table*}[!t]
\centering
\footnotesize
\setlength{\tabcolsep}{4pt}
\renewcommand{\arraystretch}{1.05}
\caption{Representative examples from the manual audit of Wason compatibility judgments.}
\begin{tabular}{p{6.5cm} p{2.2cm} p{1.3cm} p{1.3cm}}
\toprule
\textbf{Previous announcement} & \textbf{Test triple} & \textbf{Judge} & \textbf{Human} \\
\midrule
All numbers are perfect cubes & [8, 27, 64] & TRUE & TRUE \\
All three numbers are prime & [5, 17, 19] & TRUE & TRUE \\
The numbers are in strictly decreasing order & [9, 6, 3] & TRUE & TRUE \\
All numbers are prime & [2, 4, 35] & FALSE & FALSE \\
Exactly one even and two odd numbers & [4, 6, 8] & FALSE & FALSE \\
The numbers must form a strictly decreasing sequence & [1, 2, 3] & FALSE & FALSE \\
\bottomrule
\end{tabular}
\label{tab:wason-compat-judge-examples}
\end{table*}

\begin{table*}[!t]
\centering
\footnotesize
\setlength{\tabcolsep}{4pt}
\renewcommand{\arraystretch}{1.05}
\caption{Representative examples from the manual audit of Blicket announcement judgments.}
\begin{tabular}{p{1.8cm} p{2.2cm} p{5.4cm} p{1.3cm} p{1.3cm}}
\toprule
\textbf{True rule} & \textbf{True ids} & \textbf{Model announcement} & \textbf{Judge} & \textbf{Human} \\
\midrule
disjunctive & [0,1,2] & relevant=[object 0, object 1, object 2]; rule=contains at least one of object 0, 1, or 2 & TRUE & TRUE \\
conjunctive & [0,3] & relevant=[object 0, object 3]; rule=objects 0 and 3 must be placed on the device together & TRUE & TRUE \\
xor & [0,7] & relevant=[object 0, object 7]; rule=the device turns on if exactly one of object 0 or object 7 is present & TRUE & TRUE \\
conjunctive & [0,3] & relevant=[object 0, object 1, object 2, object 3]; rule=the device turns on if all objects are placed on it & FALSE & FALSE \\
xor & [0,2,3] & relevant=[object 0, object 2, object 3]; rule=the device turns on if any one of them is present & FALSE & FALSE \\
disjunctive & [0,2] & relevant=[object 0, object 3]; rule=device turns on if object 0 or object 3 is present & FALSE & FALSE \\
\bottomrule
\end{tabular}
\label{tab:blicket-guess-judge-examples}
\end{table*}

\begin{table*}[!t]
\centering
\footnotesize
\setlength{\tabcolsep}{4pt}
\renewcommand{\arraystretch}{1.05}
\caption{Representative examples from the manual audit of Blicket compatibility judgments.}
\begin{tabular}{p{8cm} p{2.2cm} p{1.3cm} p{1.3cm}}
\toprule
\textbf{Previous announcement} & \textbf{Tested object ids} & \textbf{Judge} & \textbf{Human} \\
\midrule
relevant=[object 0, object 2]; rule=must include both object 0 and object 2 & [0, 2] & TRUE & TRUE \\
relevant=[object 1, object 7]; rule=exactly one of object 1 or object 7 is required to turn the device on, but not both & [5, 7] & TRUE & TRUE \\
relevant=[object 0]; rule=object 0 alone is sufficient to turn the device on & [0] & TRUE & TRUE \\
relevant=[object 0, object 1, object 3]; rule=The device turns on when at least one of object 0, object 1, or object 3 is placed on it. & [2] & FALSE & FALSE \\
relevant=[object 0, object 2]; rule=contains object 0 or object 2 & [1, 3] & FALSE & FALSE \\
relevant=[object 0, object 1, object 2, object 3]; rule=all objects must be placed on the device & [0, 1, 3] & FALSE & FALSE \\
\bottomrule
\end{tabular}
\label{tab:blicket-compat-judge-examples}
\end{table*}

\section{Evaluation Validity Checks}

\subsection{Re-prompting Frequency Check}
\label{appendix:retry}

When a model response does not follow the required output format, we re-prompt
with the format requirement only, no task description is re-inserted.
To verify that this procedure does not introduce systematic bias, we report
re-prompting frequency on a 10\% subset of the test set alongside the task success and I:C values reported in Table~\ref{tab:replication}.

As shown in Table~\ref{tab:retry}, non-thinking models require retries on
fewer than 1\% of turns. Thinking models require retries on 36-47\% of
turns, most commonly on test turns where the model produces a rule guess
instead of a test triple after its reasoning trace. Importantly, we observe
no strong correlation between \%~Retry and either task success or I:C,
indicating that re-prompting frequency does not explain performance
differences across model-variant pairs.

\begin{table}[!t]
\centering
\caption{Re-prompting frequency alongside task success and I:C for a 10\% subset of the test set. High retry rates in thinking models reflect format errors after
reasoning traces, not task difficulty. No strong correlation is observed
between \% Retry and either performance metric.}
\label{tab:retry}
\begin{tabular}{llccc}
\toprule
\textbf{Model} & \textbf{Variant} & \textbf{Task Suc.} & \textbf{I:C} & \textbf{\% Retry} \\
\midrule
\multicolumn{5}{l}{\textit{Thinking Models}} \\
\midrule
Qwen3-8B              & Baseline   & 0.21 & 0.15 & 41.07 \\
                      & TiO        & 0.36 & 0.35 & 45.19 \\
                      & Dual-Goal  & 0.40 & 0.55 & 46.98 \\
Qwen3-14B             & Baseline   & 0.34 & 0.19 & 47.12 \\
                      & TiO        & 0.48 & 0.26 & 46.15 \\
                      & Dual-Goal  & 0.58 & 0.40 & 46.15 \\
Qwen3-32B             & Baseline   & 0.41 & 0.14 & 45.05 \\
                      & TiO        & 0.61 & 0.54 & 44.37 \\
                      & Dual-Goal  & 0.60 & 0.77 & 46.98 \\
DeepSeek-R1-LLaMA-70B & Baseline   & 0.35 & 0.32 & 37.77 \\
                      & TiO        & 0.46 & 0.48 & 39.29 \\
                      & Dual-Goal  & 0.50 & 0.61 & 36.40 \\
QwQ-32B               & Baseline   & 0.50 & 0.36 & 36.95 \\
                      & TiO        & 0.59 & 0.37 & 40.80 \\
                      & Dual-Goal  & 0.58 & 0.81 & 44.51 \\
\midrule
\multicolumn{5}{l}{\textit{Non-Thinking Models}} \\
\midrule
Qwen3-8B              & Baseline   & 0.06 & 0.35 & 0.00 \\
                      & TiO        & 0.06 & 0.30 & 0.00 \\
                      & Dual-Goal  & 0.08 & 0.57 & 0.69 \\
Qwen3-14B             & Baseline   & 0.23 & 0.72 & 0.00 \\
                      & TiO        & 0.29 & 0.90 & 0.27 \\
                      & Dual-Goal  & 0.13 & 0.41 & 0.41 \\
Qwen3-32B             & Baseline   & 0.33 & 0.50 & 0.14 \\
                      & TiO        & 0.36 & 0.63 & 0.00 \\
                      & Dual-Goal  & 0.10 & 0.46 & 0.00 \\
LLaMA-3.3-70B         & Baseline   & 0.21 & 0.37 & 0.00 \\
                      & TiO        & 0.38 & 0.50 & 0.41 \\
                      & Dual-Goal  & 0.20 & 0.53 & 0.63 \\
\bottomrule
\end{tabular}
\end{table}

\subsection{Controlled-Prior Check}
\label{appendix:controlled-prior}

A potential concern is that task success reflects alignment between a model's prior
over rules and the problem set, rather than exploration quality per se. While we note
that most evaluation benchmarks are similarly confounded by factors beyond their
primary target (e.g., answer formatting in math benchmarks, position bias in
multiple-choice settings), we nonetheless design an ablation to control for prior
effects directly.

\paragraph{Setup.} In addition to an initial triple, we provide models a candidate
set of compatible rules and instruct them that each candidate is equally likely,
thereby fixing the prior over rules to be uniform. We run 96 episodes
(8 initial triples $\times$ 12 compatible rules) using three thinking-mode base
models (Qwen3-8B, Qwen3-14B, Qwen3-32B) under Baseline and Think-in-Opposites
variants. To reflect the reduced difficulty of the setting, we limit each episode
to 9 turns (5 guess, 4 test). Results are averaged over all 8 triples and 12
rules per model-variant pair.

\begin{table}[!t]
\centering
\small
\setlength{\tabcolsep}{6pt}
\renewcommand{\arraystretch}{1.1}
\caption{Controlled-prior ablation results. Even when rule candidates are provided
with uniform probability, I:C positively correlates with task success
(Spearman $\rho = 0.341$, $p = 0.021$), and Think-in-Opposites increases
disconfirmatory testing.}
\label{tab:controlled-prior}
\begin{tabular}{llcc}
\toprule
\textbf{Model} & \textbf{Variant} & \textbf{Task Suc.} & \textbf{I:C} \\
\midrule
Qwen3-8B  & Baseline            & 68.75 & 0.551 \\
  & Think-in-Opposites  & 78.13 & 0.918 \\
Qwen3-14B & Baseline            & 79.17 & 0.871 \\
 & Think-in-Opposites  & 79.17 & 1.906 \\
Qwen3-32B & Baseline            & 77.08 & 1.000 \\
 & Think-in-Opposites  & 79.17 & 2.047 \\
\bottomrule
\end{tabular}
\end{table}

\paragraph{I:C correlates with task success under controlled prior.}
Across model-variant-triple combinations (each point averaged over 12 rules),
we observe a statistically significant positive correlation between I:C and task
success (Spearman $\rho = 0.341$, $p = 0.021$). This holds even when the prior is constant, supporting the interpretation that
I:C captures genuine exploration quality rather than prior alignment.

\paragraph{Effect of intervention under controlled prior.}
The effects of Think-in-Opposites intervention is less pronounced here compared to the main results (Tables~\ref{tab:replication}). We think this is due to the simplicity of the new task setting, leaving less room for exploration strategies to differentiate.

\subsection{Hypothesis Agreement Check}
\label{appendix:hypothesis-agreement}

Our I:C metric treats the model's announced rule as its current hypothesis. 
To test whether a single announcement reliably reflects the model's hypothesis, 
we sample 100 announcements from Qwen3-8B in thinking mode across interaction 
settings. For each announcement, we resample the model 10 times from the same 
conversation history and use Gemini-2.5-Pro to determine whether each resampled 
hypothesis agrees with the original announcement.

Across the 100 examples, the average agreement rate is 81.6\%. Moreover, for 
88 of the 100 examples, the original announcement agrees with the majority of 
the resampled hypotheses. These results suggest that the announced rule generally 
reflects the model's dominant hypothesis.

\section{Broader Impact}
\label{appendix:impact-statement}

This paper focuses on identifying and mitigating confirmation bias in LLMs. Confirmation bias may affect LLM decision-making in numerous tasks, ranging from hiring to code debugging to searching for scientific hypotheses. We urge researchers and developers who use our (or other) confirmation bias mitigation strategies to carefully test the impact across the full range of domains where their models may be used. There may be additional potential societal impacts of our work, but none that we feel must be specifically highlighted here.

\newpage
\input{checklist.tex}

\end{document}

%% file: math_commands.tex
\usepackage{amsmath,amsfonts,bm}

\def\eqref#1{equation~\ref{#1}}

\def\1{\bm{1}}

\DeclareMathAlphabet{\mathsfit}{\encodingdefault}{\sfdefault}{m}{sl}
\SetMathAlphabet{\mathsfit}{bold}{\encodingdefault}{\sfdefault}{bx}{n}



%% file: checklist.tex
\section*{NeurIPS Paper Checklist}

\begin{enumerate}

\item {\bf Claims}
    \item[] Question: Do the main claims made in the abstract and introduction accurately reflect the paper's contributions and scope?
    \item[] Answer: \answerYes{} 
    \item[] Justification: The abstract and introduction accurately state the paper’s contributions: (1) a framework and dataset for measuring confirmation bias in LLMs via the Wason 2-4-6 rule-discovery task, (2) evidence that LLMs exhibit confirmation bias that negatively correlates with task success, (3) that human-derived prompting interventions (Dual-Goal, Think-in-Opposites) reduce confirmation bias and improve performance, and (4) that distillation can internalize these behaviors, with generalization to the Blicket test and beyond rule discovery to mathematical self-correction. The claims are supported by the experimental sections on Wason-style rule discovery, intervention results, distillation, Blicket generalization, and mathematical reasoning.

\item {\bf Limitations}
    \item[] Question: Does the paper discuss the limitations of the work performed by the authors?
    \item[] Answer: \answerYes{} 
    \item[] Justification: We discuss limitations in the Limitations and Future Work
paragraph of the Conclusion. It notes that the framework primarily studies confirmation bias in controlled rule-discovery settings and, while the benefits of distillation extend beyond rule discovery to mathematical self-correction, future work is needed to evaluate whether similar improvements generalize to other real-world settings.

\item {\bf Theory assumptions and proofs}
    \item[] Question: For each theoretical result, does the paper provide the full set of assumptions and a complete (and correct) proof?
    \item[] Answer: \answerNA{} 
    \item[] Justification: The paper does not include theoretical results or formal proofs. All contributions are empirical.

    \item {\bf Experimental result reproducibility}
    \item[] Question: Does the paper fully disclose all the information needed to reproduce the main experimental results of the paper to the extent that it affects the main claims and/or conclusions of the paper (regardless of whether the code and data are provided or not)?
    \item[] Answer: \answerYes{} 
    \item[] Justification: The paper provides detailed descriptions of the dataset construction process (Appendix~\ref{appendix:data}), all prompts used for the Baseline, Think-in-Opposites, and Dual-Goal conditions (Appendices~\ref{appendix:baseline-prompt}--\ref{appendix:dualgoal-prompt}), LLM-as-a-judge prompts (Appendix~\ref{appendix:judge}), distillation hyperparameters and learning-rate sweeps (Appendix~\ref{appendix:distillation-details}), data splits, and statistical testing procedures (Appendix~\ref{appendix:intervention-stats}).

\item {\bf Open access to data and code}
    \item[] Question: Does the paper provide open access to the data and code, with sufficient instructions to faithfully reproduce the main experimental results, as described in supplemental material?
    \item[] Answer: \answerYes{} 
    \item[] Justification: The repository linked on page~1 contains the datasets,
evaluation and distillation code, configuration files, and full model
outputs, along with instructions for reproducing the main experimental
results.

\item {\bf Experimental setting/details}
    \item[] Question: Does the paper specify all the training and test details (e.g., data splits, hyperparameters, how they were chosen, type of optimizer) necessary to understand the results?
    \item[] Answer: \answerYes{} 
    \item[] Justification: Data splits (4 train / 2 validation / 4 test rule groups), evaluation metrics, and the interaction protocol are described in Section~\ref{sec:framework} and Appendix~\ref{appendix:data}. Appendix~\ref{appendix:distillation-details} provides full distillation hyperparameters (LoRA rank, learning rate sweeps, batch size, epochs, warmup ratio, and checkpoint selection).

\item {\bf Experiment statistical significance}
    \item[] Question: Does the paper report error bars suitably and correctly defined or other appropriate information about the statistical significance of the experiments?
    \item[] Answer: \answerYes{} 
    \item[] Justification: The paper reports statistical significance for all main comparisons using paired permutation tests (50,000 permutations) with Benjamini--Hochberg correction for multiple comparisons, detailed in Appendices~\ref{appendix:intervention-stats} and~\ref{appendix:distillation-stats}. Figures~\ref{fig:ic_correlations} and~\ref{fig:ic_correlations_blickets} display 95\% confidence intervals.

\item {\bf Experiments compute resources}
    \item[] Question: For each experiment, does the paper provide sufficient information on the computer resources (type of compute workers, memory, time of execution) needed to reproduce the experiments?
    \item[] Answer: \answerYes{} 
    \item[] Justification: Compute details are provided in Appendix~\ref{appendix:distillation-details}.
Fine-tuning was conducted on H200 GPUs (140GB each): 1 GPU for
Qwen3-8B students and 2 GPUs for Qwen3-32B students, with training
runs taking 2 to 4 days depending on validation loss convergence.
    
\item {\bf Code of ethics}
    \item[] Question: Does the research conducted in the paper conform, in every respect, with the NeurIPS Code of Ethics \url{https://neurips.cc/public/EthicsGuidelines}?
    \item[] Answer: \answerYes{} 
    \item[] Justification: The work evaluates and mitigates confirmation bias in LLM reasoning using synthetic tasks and existing model APIs/checkpoints. To the best of the authors' knowledge, the research conforms to the NeurIPS Code of Ethics.

\item {\bf Broader impacts}
    \item[] Question: Does the paper discuss both potential positive societal impacts and negative societal impacts of the work performed?
    \item[] Answer: \answerYes{} 
    \item[] Justification: We discuss broader impact in Appendix ~\ref{appendix:impact-statement}. This discusses how our strategies can reduce confirmation bias and affect LLM decision-making in numerous tasks, ranging from hiring to code debugging to searching for scientific hypotheses.
    
\item {\bf Safeguards}
    \item[] Question: Does the paper describe safeguards that have been put in place for responsible release of data or models that have a high risk for misuse (e.g., pre-trained language models, image generators, or scraped datasets)?
    \item[] Answer: \answerNA{} 
    \item[] Justification: The paper does not release a high-risk pretrained language model, scraped dataset, or other asset with clear misuse risk. The datasets are controlled synthetic rule-discovery tasks rather than sensitive real-world data.

\item {\bf Licenses for existing assets}
    \item[] Question: Are the creators or original owners of assets (e.g., code, data, models), used in the paper, properly credited and are the license and terms of use explicitly mentioned and properly respected?
    \item[] Answer: \answerYes{} 
    \item[] Justification: All datasets used are cited: the Large Dataset of Generalization Patterns in the Number Game~\citep{DVN/A8ZWLF_2018} and Tenenbaum's Number Game~\citep{Tenenbaum2000RulesSimilarity} are credited for human-derived rules. The Blicket test environment is credited to~\citet{gxchen2025language}. AIME 2025~\citep{opencompass2025aime} and Beyond AIME~\citep{bytedance_seed_2025_beyondaime} are cited to their respective dataset sources. All LLMs are cited with their original papers.

\item {\bf New assets}
    \item[] Question: Are new assets introduced in the paper well documented and is the documentation provided alongside the assets?
    \item[] Answer: \answerYes{} 
    \item[] Justification: The new Wason rule-discovery dataset and Blicket evaluation dataset are fully documented in Appendices~\ref{appendix:data} and~\ref{appendix:blicket-dataset}, including generation procedures, filtering criteria, data splits, and all rule definitions and test instances (Tables~\ref{tab:rules-train}--\ref{tab:instances-test}). The full datasets and accompanying code are publicly
available in the repository linked on page~1.

\item {\bf Crowdsourcing and research with human subjects}
    \item[] Question: For crowdsourcing experiments and research with human subjects, does the paper include the full text of instructions given to participants and screenshots, if applicable, as well as details about compensation (if any)? 
    \item[] Answer: \answerNA{} 
    \item[] Justification: The paper does not involve crowdsourcing or human subjects. All experiments are conducted with LLMs.

\item {\bf Institutional review board (IRB) approvals or equivalent for research with human subjects}
    \item[] Question: Does the paper describe potential risks incurred by study participants, whether such risks were disclosed to the subjects, and whether Institutional Review Board (IRB) approvals (or an equivalent approval/review based on the requirements of your country or institution) were obtained?
    \item[] Answer: \answerNA{} 
    \item[] Justification: The paper does not involve human subjects; IRB approval is not applicable.

\item {\bf Declaration of LLM usage}
    \item[] Question: Does the paper describe the usage of LLMs if it is an important, original, or non-standard component of the core methods in this research? Note that if the LLM is used only for writing, editing, or formatting purposes and does \emph{not} impact the core methodology, scientific rigor, or originality of the research, declaration is not required.
    \item[] Answer: \answerYes{} 
    \item[] Justification: LLMs are a central component of the methodology. They are used as (1) agents being evaluated for confirmation bias, (2) LLM-as-a-judge for assessing rule correctness and triple compatibility (Appendix~\ref{appendix:judge}), (3) a data generation tool for synthetic rule sets (Appendix~\ref{appendix:ruleprompt}), and (4) teacher models for knowledge distillation (Section~\ref{sec:distillation}). All LLM roles are described in detail in the paper.

\end{enumerate}